\documentclass{article}

\usepackage{microtype}
\usepackage{graphicx}
\usepackage{subfigure}
\usepackage{booktabs} 

\usepackage{hyperref}
\usepackage{xspace}
\usepackage{enumitem}

\usepackage[accepted]{icml2025}

\usepackage{amsmath, bm}
\usepackage{amssymb}
\usepackage{mathtools}
\usepackage{amsthm}

\usepackage{graphicx}      
\usepackage{subcaption}    

\usepackage[capitalize,noabbrev]{cleveref}

\theoremstyle{plain}
\newtheorem{theorem}{Theorem}[section]

\theoremstyle{definition}
\newtheorem{definition}[theorem]{Definition}

\theoremstyle{remark}

\usepackage[textsize=tiny]{todonotes}

\newcommand{\ie}{\textit{i}.\textit{e}.}
\newcommand{\eg}{\textit{e}.\textit{g}.}

\def\vepsilon{{\bm{\epsilon}}}
\def\vx{{\bm{x}}}
\def\vy{{\bm{y}}}

\def\vzero{{\bm{0}}}

\def\vw{{\bm{w}}}

\def\vmu{{\bm{\mu}}}

\def\mI{{\bm{I}}}

\def\mW{{\bm{W}}}

\def\S{{S}}

\def\SK{\S_K}

\def\Skmx{\S_{k-1} \oplus x^{(k)}}

\newtheorem{asu}{Assumption}

\newcounter{subassumption}[asu]
\renewcommand{\thesubassumption}{(\textit{\roman{subassumption}})}
\makeatletter
\renewcommand{\p@subassumption}{\theasu}
\makeatother

\renewcommand{\thesubassumption}{(\alph{subassumption})}

\newcommand{\subasu}{%
  \refstepcounter{subassumption}%
  \thesubassumption~\ignorespaces}

\def\mytitle{In-Context Learning with Hypothesis-Class Guidance}

\def\hz{\textcolor[RGB]{0,146,50}{$z$}\xspace}

\usepackage[most]{tcolorbox}
\newtcolorbox{highlight}[1][]{
    enhanced,
    colback=yellow!10,
    colframe=gray!30,
    boxrule=0.5pt,
    arc=2pt,
    leftrule=3pt,
    rightrule=3pt,
    toprule=1pt,
    bottomrule=1pt,
    breakable,
    #1
}

\icmltitlerunning{\mytitle}

\begin{document}

\twocolumn[
\icmltitle{\mytitle}



\icmlsetsymbol{equal}{*}

\begin{icmlauthorlist}
\icmlauthor{Ziqian Lin}{xxx}
\icmlauthor{Shubham Kumar Bharti}{xxx}
\icmlauthor{Kangwook Lee}{zzz}
\end{icmlauthorlist}

\icmlaffiliation{xxx}{Department of Computer Science, University of Wisconsin-Madison, Madison, Wisconsin, USA}
\icmlaffiliation{zzz}{Department of Electrical \& Computer Engineering, University of Wisconsin-Madison, Madison, Wisconsin, USA}

\icmlcorrespondingauthor{Kangwook Lee}{kangwook.lee@wisc.edu}

\icmlkeywords{Machine Learning, ICML}

\vskip 0.3in
]



\printAffiliationsAndNotice{}  

\begin{abstract}
Recent research has investigated the underlying mechanisms of in-context learning (ICL) both theoretically and empirically, often using data generated from simple function classes.
However, the existing work often focuses on the sequence consisting solely of labeled examples, while in practice, labeled examples are typically accompanied by an \emph{instruction}, providing some side information about the task. 
In this work, we propose \emph{ICL with hypothesis-class guidance (ICL-HCG)}, a novel synthetic data model for ICL where the input context consists of the literal description of a (finite) hypothesis class $\mathcal{H}$ and $(x,y)$ pairs from a hypothesis chosen from $\mathcal{H}$.
Under our framework ICL-HCG, we conduct extensive experiments to explore: 
(i) a variety of generalization abilities to new hypothesis classes; 
(ii) different model architectures;
(iii) sample complexity;
(iv) in-context data imbalance;
(v) the role of instruction; and
(vi) the effect of pretraining hypothesis diversity.
As a result, we show that 
(a) Transformers can successfully learn ICL-HCG and generalize to unseen hypotheses and unseen hypothesis classes, and (b) compared with ICL without instruction, ICL-HCG achieves significantly higher accuracy, demonstrating the role of instructions. 
The code is available at:  
\url{https://github.com/UW-Madison-Lee-Lab/ICL-HCG}.
\end{abstract}

\section{Introduction}
\paragraph{LLMs and ICL}
Large Language Models (LLMs)~\citep{zhao2023survey} have garnered widespread attention for their ability to solve complex tasks using simple text prompts.
Among their many capabilities, in-context learning (ICL)~\citep{NEURIPS2020_1457c0d6} is particularly striking.
ICL enables LLMs to adapt to new tasks by conditioning on provided examples, effectively allowing them to learn from context without explicit parameter updates.
Understanding how such behavior emerges in LLMs remains an intriguing and challenging problem.
\begin{figure*}[th!]
    \centering
    \includegraphics[width = 1.0\textwidth]{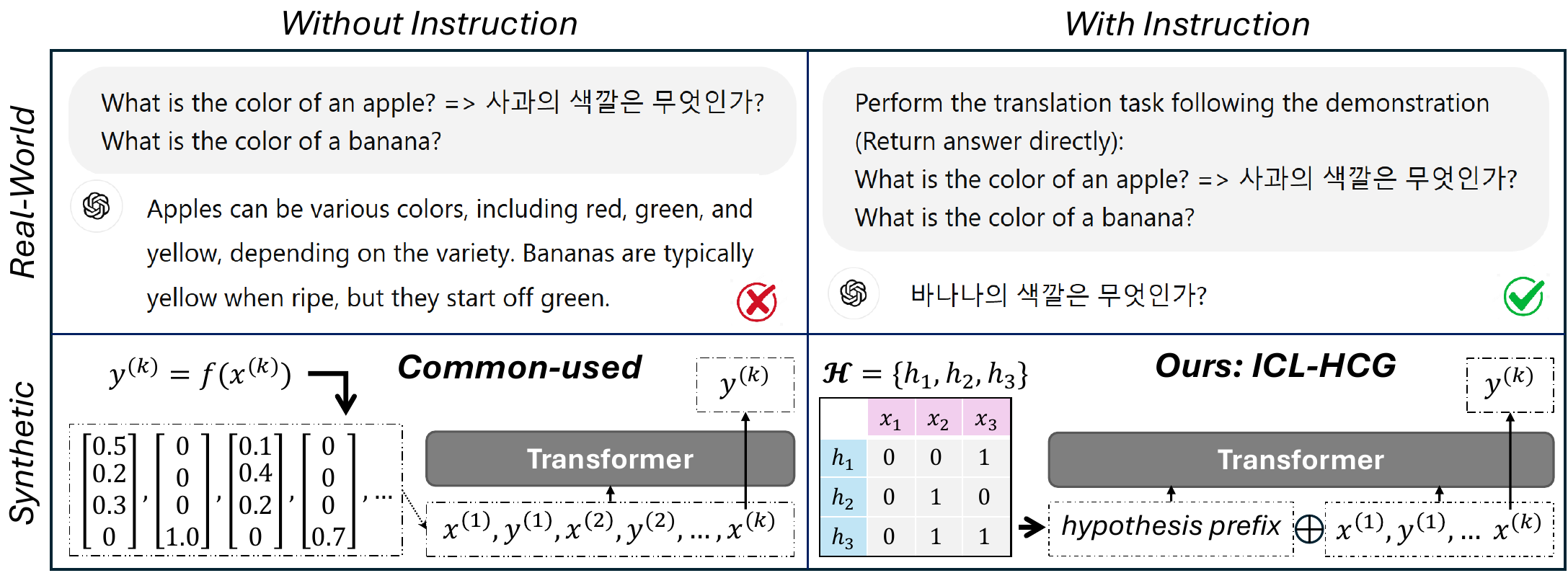}
    \caption{\textbf{Common ICL framework \textit{vs.} ours.}  
    Conventional frameworks with synthetic datasets often construct sequences by concatenating multiple \((\vx, \vy)\) pairs, overlooking the importance of instructions.
    In contrast, our approach explicitly incorporates instructions through a \textit{hypothesis prefix}. Specifically, we transform the hypothesis class \(\mathcal{H}\) into a sequence that is prepended to the sequence of \((\vx, \vy)\) pairs and then fed into a Transformer. We refer to this method as \textit{in-context learning with hypothesis-class guidance (ICL-HCG)}.
    (Real-world examples are demonstrated using the GPT-4 Legacy model.)
    }
    \label{fig:main}
\end{figure*}

\paragraph{Existing Efforts for Understanding ICL} To elucidate the mechanisms behind ICL, researchers have
constructed a variety of synthetic datasets~\citep{garg2022can,li2023transformers,BaiCWXM23}.
These datasets
typically involve sequences consisting of input-output pairs
\(\{(\vx^{(i)}, y^{(i)})\}\), where each output
\(y^{(i)}\) is generated by a simple underlying function
\(f(\vx^{(i)})\). For example, \citet{garg2022can} focus on noiseless
linear regression, where each input is sampled from an
isotropic Gaussian by $\vx^{(i)} \sim \mathcal{N}(\vzero, \mI_d)$, and the corresponding
output is given by \(y^{(i)} = \langle \vx^{(i)}, \vw \rangle\) with
\(\vw \sim \mathcal{N}(\vzero, \mI_d)\) for each sequence.
During training and inference, the model
receives a sequence consisting of \(k\) demonstration pairs
\((\vx^{(1)}, y^{(1)}, \dots, \vx^{(k-1)}, y^{(k-1)})\) followed
by a query input \(\vx_{\text{query}}\). This setup allows the model to
infer the correct response for \(\vx_{\text{query}}\) conditioned on in-context examples.
Various extensions have been proposed, including using Gaussian mixtures rather than a single
Gaussian for task priors~\citep{lin2024dual}, employing non-linear
functions~\citep{BaiCWXM23}, and introducing multiple intermediate
``chain-of-thought''~\citep{wei2022chain} steps within each $(\vx,\vy)$ pair~\citep{li2024how}.

\textbf{Motivation}
While a variety of data models have been studied to advance our understanding of ICL, a gap remains between these datasets and real-world ICL scenarios.
In practice, users often provide LLMs with \emph{an instruction} in addition to labeled demonstrations, containing the descriptions of the task in mind. 
See Fig.~\ref{fig:main} for the visualization.
The top-left and top-right panels show experimental results using the GPT-4 Legacy model, highlighting the effect of instruction.
In the top-left panel, the user provides a one-shot English-Korean translation pair without specifying the instruction, leading to an incorrect translation.
In contrast, the top-right panel includes the instruction—\emph{“perform the translation task following the demonstration”}—guiding the model to produce a correct translation, emphasizing the importance of the task descriptions.
In fact, instructions are known to enhance the accuracy of ICL in general~\citep{NEURIPS2020_1457c0d6}.
However, most existing synthetic data frameworks overlook this crucial aspect, neglecting the role of instructions in guiding the learning process.
Motivated by this limitation, we ask:
\begin{center}
    \it Can we design a synthetic data framework for ICL that better captures the practical use scenarios of ICL by incorporating both instructions and labeled samples?
\end{center}
Notably, two recent works~\citep{xuanyuan2024on,huang2024task} adopt prefix as instruction to implicitly provide information on the task.
In contrast, our approach explicitly provides a hypothesis class as a prefix to the Transformer, guiding the model's understanding of the intended task.

\paragraph{Our Synthetic Data Model}  
We propose a novel synthetic data model, \textit{in-context learning with hypothesis-class guidance (ICL-HCG)}, illustrated in the bottom-right panel of Fig.~\ref{fig:main}, which integrates a hypothesis class into the ICL procedure.  
Specifically, besides the usual sequences of $(\vx,\vy)$ pairs, a hypothesis class is embedded as a hypothesis prefix and fed into the Transformer (more details in Fig.~\ref{fig:framework} of Sec.~\ref{subsec:framework}). 
Leveraging this framework, we explore several aspects of Transformer behavior on the ICL-HCG task:
(i) We evaluate the generalization ability of trained models to new hypothesis classes, new hypotheses, and various sizes of hypothesis classes;
(ii) We compare different model architectures (Transformer, Mamba, LSTM, and GRU), highlighting their distinct properties on these generalizations;
(iii) We examine the sample complexity required for achieving ID and OOD hypothesis class generalization and discover that merely a few dozen training hypothesis classes are sufficient for near-perfect generalization.
(iv) We examine the effect of imbalanced in-context samples, demonstrating that imbalance can slow down the training process;
(v) We assess the benefit of incorporating a hypothesis prefix, which notably enhances the accuracy of ICL;
(vi) We show pretraining hypothesis diversity can significantly improve the accuracy of ICL when with instruction.

We summarize our contributions as follows:
\begin{itemize}[leftmargin=0.2cm]
    \item We propose a novel synthetic data model, namely in-context learning with hypothesis-class guidance (ICL-HCG) that integrates a hypothesis class into the ICL procedure.
    This design provides a controlled testbed for diverse experiments to study behaviors of ICL with instruction.
    \item We perform extensive empirical evaluations on our framework.
    Most interestingly, we demonstrate that (a) Transformers can successfully learn ICL-HCG and such a learned ability can generalize to unseen hypotheses and unseen hypothesis classes, and (b) compared with ICL without instruction, ICL-HCG achieves significantly higher accuracy on ICL, demonstrating the role of instructions.
\end{itemize}


\section{Related Works}
\label{sec:RelatedWork}


\citet{garg2022can} first construct synthetic data with pretraining and testing sequences generated from noiseless linear regression tasks.
Specifically, \citet{garg2022can} sample all input vectors \(\vx\) from an isotropic Gaussian distribution \(\mathcal{N}(\vzero, \mathbf{I})\). Within each sequence, the outputs are given by
\(
y^{(i)} = \langle \vw, \vx^{(i)} \rangle,
\)
where \(\vw\) is drawn from \(\mathcal{N}(\vzero, \mathbf{I}_d)\). 
For ICL inference, the prompt takes the form
\(
(\vx^{(1)}, y^{(1)}, \vx^{(2)}, y^{(2)}, \ldots, \vx^{(k-1)}, y^{(k-1)}, \vx_{\text{query}})
\)
where the \(k-1\) pairs \(\{(\vx^{(i)}, y^{(i)})\}_{i=1}^{k-1}\) serve as demonstrations illustrating the linear relationship governed by \(\vw\).
The model then predicts the output for \(\vx_{\text{query}}\).

\paragraph{Noiseless Linear Regression} Based on the well-defined problem setup by~\citet{garg2022can} using noiseless linear regression, researchers systematically study the mechanisms of ICL and properties of Transformer.
For instance, there is a particular interesting line of research on connecting ICL to gradient descent, firstly hinted by~\citet{garg2022can}.
\citet{akyurek2022learning} and \citet{von2022transformers} then show that one attention layer can be exactly constructed to perform gradient descent, and empirically find similarities between ICL and gradient descent. 
Further, \citet{ahn2023transformers} theoretically show that under certain conditions, Transformers trained on noiseless linear regression tasks minimizing the pretraining loss will implement gradient descent algorithm.
Nevertheless, \citet{fu2024transformers} show that Transformers learn to approximate second-order optimization methods for ICL, sharing a similar convergence rate as Iterative Newton’s Method.
Besides gradient descent, there are lots of other interesting topics on ICL and Transformers based on this linear regression setting, such as looped Transformer~\citep{Yang0NP24,GatmirySRJK24}, training dynamic~\citep{zhang2024trained, HuangCL24, kim2024transformers}, generalization~\citep{panwar2024incontext}, etc.

\paragraph{Noisy Linear Regression} Such a simple noiseless linear regression task is further extended to variants.
By extending the linear regression to noisy linear regression---$y=\langle \vx, \vw \rangle + \epsilon$,
\citet{li2023transformers} analyze the generalization and stability of ICL. 
\citet{WuZCBGB24} and \citet{raventos2024pretraining} analyze the effect of task diversity on the attention model's ICL risk.
Via extending the regression tasks sampling from Gaussian to Gaussian mixture, \citet{lin2024dual} show ICL exhibits two different modes including task retrieval and learning.
With the tasks of $\vy=\mW\vx+\vepsilon$ where $\mW$ is a matrix rather than a vector, \citet{ChenSWY24} examine the training dynamic of multi-head attention for ICL.

\paragraph{More than Linear Regression} Beyond linear regression, researchers are also interested in non-linear regression and classification.
The research directions are scattered, and we list them as follows.
\citet{BaiCWXM23} show that Transformers can perform in-context algorithm selection, \ie, adaptively selecting different ICL algorithms such as gradient descent, least square, or ridge regression.
\citet{BhattamishraPBK24} show Transformer can learn a variety of Boolean function classes.
\citet{cheng2024transformers} provide evidence that Transformers can learn to implement gradient descent to enable them to learn non-linear functions.
\citet{0004HMWXS024} show that trained Transformer achieves near-optimal ICL performance under $y=\langle \vw, f(\vx)\rangle$, where $f$ is a shallow neural network (MLP).
Examining linear and non-linear regression tasks, \citet{fan2024transformers} and~\citet{tripuraneni2024can} show Transformer can perform ICL on composited or mixed tasks of pretrained linear or non-linear tasks, and \citet{yadlowsky2024can} examine whether trained Transformers can generalize to new tasks beyond pretraining.
\citet{park2024can} examine whether Mamba can in-context learn a variety of synthetic tasks.
Via examining regression and classification tasks, \citet{kim2024task} show task diversity helps shorten the ICL plateau pretraining.
\citet{rameshcompositional} assume there are multiple functions composited to connect $\vx$ and $\vy$ pair, \eg, $\vy=f_1\circ f_2\circ f_3(\vx)$ to study the compositional capabilities of Trasnformer.
\citet{li2024nonlinear} study how non-linear Transformer learns binary classification.

\paragraph{Synthetic Dataset with Instruction}
To the best of our knowledge, there are two articles on synthetic datasets with instructions.
\citet{huang2024task} append an additional vector $\vmu$ to the sequences with interleaved input-output format, which leads to the sequence $(\vmu, \vx^{(1)}, \vw^\top\vx^{(1)}, \vx^{(2)}, \vw^\top\vx^{(2)}, \ldots)$ in which $\vx^{(i)}\sim\mathcal{N}(\vmu,\mI)$, and show that the trained Transformer can achieve significantly lower loss on ICL when the task descriptor $\vmu$ is provided.
The work of \citet{xuanyuan2024on} is most closely related to us.
It develops a new synthetic dataset based on task $((a\cdot x)\circ(b\cdot y))\mod p=r$, where $(x,y)$ is the input, $r$ is the output, $\circ$ is an operation ($+,-,/$), and each task is defined by the parameters $(a,b,\circ)$ ($p$ is a constant).
The instruction is constructed as $(a_l,a_u,b_l,b_u,\circ)$, where $a_l$ and $a_u$ are the lower and upper bounds of $a$ (similar for $b$), and $\circ$ is the operation.
Therefore, the instruction constrains the possible tasks, \ie, provide information on the underlying true task of in-context samples.
With such a setting, \citet{xuanyuan2024on} study how the information provided by instruction affects the ICL accuracy.

\section{Meta-Learning for ICL-HCG}
Training a learner to perform ICL aligns with the concept of meta-learning, as it enables adaptation to new tasks using in-context examples. While prior studies~\citep{garg2022can,fan2024transformers,raventos2024pretraining} train Transformers for ICL on sequences of the form \((x_1, y_1, x_2, y_2, \dots, x_k, y_k)\) without explicit instructions, our work investigates whether a Transformer trained for ICL with instructions, namely ICL-HCG, can generalize to new ICL-HCG tasks.


\subsection{Two Types of Tasks in ICL-HCG}
\label{sec:problem-definition}
We consider two types of tasks in ICL-HCG, both constructed from a finite hypothesis class
\(
    \mathcal{H} = \{h^{(1)}, h^{(2)}, \ldots, h^{|\mathcal{H}|}\}
\)
over a finite input space
\(
    \mathcal{X} = \{x_1, x_2, \ldots, x_{|\mathcal{X}|}\}
\)
and a binary output space
\(
    \mathcal{Y} = \{0, 1\}.
\)

\paragraph{Label prediction}
Consider a hypothesis class \(\mathcal{H}\) and a sequence consisting of training data and a test point
\[
    \Skmx = (x^{(1)}, y^{(1)}, \ldots, x^{(k-1)}, y^{(k-1)}, x^{(k)})
\]
where for all $i$, \(y^{(i)} = h(x^{(i)})\) for a specific \(h \in \mathcal{H}\), and \(x^{(k)}\) is a test query input. The objective is to predict the label 
\[
    y^{(k)} = h\bigl(x^{(k)}\bigr).
\]
We refer to this as label prediction, with input-output pairs:
\[
    i_{\text{I},k} = \bigl(\mathcal{H}, \Skmx\bigr),
    \quad
    o_{\text{I},k} = y^{(k)}.
\]

\paragraph{Hypothesis identification}
Given a hypothesis class \(\mathcal{H}\) and a sequence (namely \emph{ICL sequence}) 
\[
    \SK = (x^{(1)}, y^{(1)}, \ldots, x^{(K)}, y^{(K)}),
\]
where for all $i$, \(y^{(i)} = h(x^{(i)})\) for a specific  \(h \in \mathcal{H}\),
the goal is to identify the underlying hypothesis \(h\).
Denote this as hypothesis identification, with:
\[
    i_{\text{II},K} = \bigl(\mathcal{H}, \SK \bigr),
    \quad
    o_{\text{II},K} = h.
\]

\paragraph{Meta-learning}
Label prediction uses \(k-1\) samples to predict the label of a new query \(x^{(k)}\), while hypothesis identification directly outputs \(h\).
Both label prediction and hypothesis identification can be viewed as attempts to identify \(h\) from $\mathcal{H}$ via empirical risk minimization (ERM) using the dataset 
\(\{(x^{(i)},y^{(i)})\}\).
Our meta-learning aims at learning to do ERM for different hypothesis classes when these hypothesis classes are given as input along with $(x,y)$ pairs.

\subsection{Sample Generation}
We consider the following two approaches for generating samples of ICL-HCG tasks.
\begin{asu}[i.i.d.\ Generation]
\label{asu:iid}
Given hypothesis classes
\(\{\mathcal{H}_i\}_{i=1}^{N}\), input space \(\mathcal{X}\), and an integer \(K\):\\
\subasu\label{asu:setting1}Sample a hypothesis class \(\mathcal{H}\) from \(\{\mathcal{H}_i\}_{i=1}^{N^\text{train}}\);\\
\subasu\label{asu:setting2}Sample a hypothesis \(h\) uniformly at random from \(\mathcal{H}\);\\
\subasu\label{asu:setting3}Sample \(K\) inputs \(\{x^{(i)}\}_{i=1}^{K}\) i.i.d.\ from \(\mathrm{Uniform}(\mathcal{X})\);\\
\subasu\label{asu:setting4}Generate \(y^{(i)} = h(x^{(i)})\) for each \(i \in [K]\);\\
\subasu\label{asu:setting5}\(\Skmx = [x^{(1)},y^{(1)}, \ldots, x^{(k)}]\) for label prediction;\\
\subasu\label{asu:setting6}{\tiny \(\SK = [x^{(1)},y^{(1)}, \ldots, x^{(K)},y^{(K)}]\)} for hypothesis identification.\\
\end{asu}

\begin{asu}[Opt-T Generation]
\label{asu:optt}
Given hypothesis classes
\(\{\mathcal{H}_i\}_{i=1}^{N}\), input space \(\mathcal{X}\), and an integer \(K\):\\
\subasu Sample a hypothesis class \(\mathcal{H}\) from \(\{\mathcal{H}_i\}_{i=1}^{N^\text{test}}\);\\
\subasu Sample a hypothesis \(h\) uniformly randomly from \(\mathcal{H}\);\\
\subasu Construct \emph{optimal teaching set}\footnote{The optimal teaching set~\citep{zhu2015machine} is the smallest set of \((x,y)\) pairs that uniquely identifies \(h\) among all candidates in \(\mathcal{H}\).} of \(h\) with respect to \(\mathcal{H}\);\\
\subasu Randomly duplicate elements from this optimal teaching set until its size reaches \(K\). Assign indices \(1\) through \(K\) arbitrarily to these \((x,y)\) pairs;\\
\subasu {\tiny \(\SK = [x^{(1)},y^{(1)}, \ldots, x^{(K)},y^{(K)}]\)} for hypothesis identification.
\end{asu}

\subsection{Meta Training and Testing}
\paragraph{Training}
Given a set of training hypothesis classes $\{\mathcal{H}_i^\text{train}\}_{i=1}^{N^\text{train}}$, the meta-learner is trained in a multi-task setting to minimize the following loss:
\begin{align}
\label{eq:loss}
    \mathcal{L} = \mathcal{L}_1(f_\theta(i_{\text{II},K}), o_{\text{II},K}) + \sum_{k=1}^K \mathcal{L}_2(f_\theta(i_{\text{I},k}), o_{\text{I},k}),
\end{align}
where we generate $\mathcal{H}$, $h$, and $\SK$ following \hyperref[asu:iid]{i.i.d. Generation}, inherently defining $(i_{\text{II},K}, o_{\text{II},K})$ and $(i_{\text{I},k}, o_{\text{I},k})$.
The loss is indeed implemented with additional terms, and we will further clarify the loss in Sec.~\ref{subsec:framework}, Eq.~\ref{eq:lossTF}.

\paragraph{Testing}
Given a set of testing hypothesis classes $\{\mathcal{H}_i^\text{test}\}_{i=1}^{N^\text{test}}$, we consider two types of testing.
\begin{itemize}[topsep=0.1em, partopsep=0em, leftmargin=*]
    \item \textbf{Label prediction}: We generate $(i_{\text{I},k}, o_{\text{I},k})$ following \hyperref[asu:iid]{i.i.d. Generation}, and then measure whether the learner $f$ predict $f(i_{\text{I},k})$ correctly for each $k\in[K]$;
    \item \textbf{Hypothesis identification}: We generate $(i_{\text{II},K}, o_{\text{II},K})$ using \hyperref[asu:optt]{Opt-T Generation} and evaluate whether the learner $f$ predicts $f(i_{\text{II}})$ correctly.
    This setting tests whether the learner acquires the ability to identify the underlying hypothesis with minimal information.
\end{itemize}



\subsection{Four Types of Generalization}
\paragraph{Hypothesis universe $\mathcal{H}^{\text{uni}}$}
Given an input space 
$
    \mathcal{X} = \{x_1, x_2, \ldots, x_{|\mathcal{X}|}\}
$
and a binary output space
$
    \mathcal{Y} = \{0, 1\},
$
We define the hypothesis universe $\mathcal{H}^{\text{uni}}=\mathcal{Y}^{\mathcal{X}}$ as the collection of all possible binary classification hypotheses.
This universe contains $M=2^{|\mathcal{X}|}$ distinct hypotheses, serving as a hypothesis pool to constructing training and testing hypothesis classes.

\begin{figure}[h!]
    \centering
    \includegraphics[width = 0.5\textwidth]{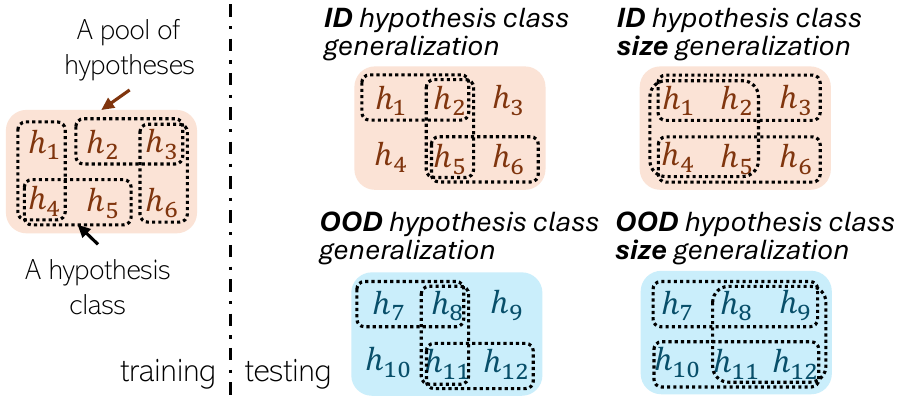}
    \caption{\textbf{Four types of generalization.}
    An illustration of the four types of generalization.}
    \label{fig:framework}
\end{figure}

In meta-learning, the goal is to train a model that is able to rapidly adapt to new tasks.  
Testing on new tasks can be considered as measuring the OOD generalization.
Under our ICL-HCG framework, we consider four types of OOD generalizations.
First, we examine whether the learner generalizes to a new testing hypothesis class (the hypothesis class is unseen during training) that may or may not contain hypotheses seen during training, referred to as in-distribution (ID) and out-of-distribution (OOD) hypothesis class generalization, respectively.

\begin{definition}[ID Hypothesis Class Generalization]
\label{def:SpaceGeneralization}
Given $\mathcal{H}^{\mathrm{uni}}$ of size $M$, we enumerate all $C(M, m) = \frac{M!}{m!(M - m)!}$ distinct hypothesis classes, each containing $m$ 
hypotheses.
We then randomly \emph{subsample} these classes into disjoint training and testing subsets, ensuring that no testing hypothesis class appears in the training set (although individual hypotheses may overlap).
By training on randomly selected training hypothesis classes and evaluating on unseen testing hypothesis classes, we assess generalization to new hypothesis classes consisting of ID hypotheses.
\end{definition}

\begin{definition}[OOD Hypothesis Class Generalization]
\label{def:HypothesisGeneralization}
Given $\mathcal{H}^{\text{uni}}$ of size $M$, 
we partition it into disjoint training and testing subsets of sizes 
$M^\text{ID}$ and $M^\text{OOD}$, respectively. 
We then generate training hypothesis classes from $M^\text{ID}$ and testing hypothesis classes from $M^\text{OOD}$, each containing $m$ hypotheses. 
We train the learner on the training hypothesis classes and evaluate on the testing hypothesis classes. 
Because no testing hypothesis appears during training, 
this setup probes how well the learner generalizes
to entirely new hypotheses, \ie, OOD hypotheses.
\end{definition}

We then consider whether the learner can generalize to hypothesis classes of various sizes. Building on the concepts of ID and OOD hypothesis class generalization, we introduce size generalizations as follows.

\begin{definition}[ID Hypothesis Class Size Generalization]
\label{def:Space&SizeGeneralization}
Building on the setting of ID hypothesis class generalization, while maintaining non-identical training and testing hypothesis classes, we allow training hypothesis class to include various number of hypotheses $m\in\mathcal{M}\subsetneqq[L]$.
We investigate whether the learner can perform well on hypothesis classes with other sizes $m\in [L]\setminus\mathcal{M}$, where $[L]=\{1,2,\ldots,L\}$.
\end{definition}

\begin{definition}[OOD Hypothesis Class Size Generalization]
\label{def:Hypothesis&SizeGeneralization}
Based on the setting of OOD hypothesis class generalization, while maintaining non-identical training and testing hypotheses, we allow training hypothesis class to include various number of hypotheses $m\in\mathcal{M}\subsetneqq[L]$.
We investigate whether the learner can perform well on hypothesis classes with various sizes $m\in [L]\setminus\mathcal{M}$, where $[L]=\{1,2,\ldots,L\}$.
\end{definition}

\begin{figure}[h!]
    \centering
    \includegraphics[width = 0.475\textwidth]{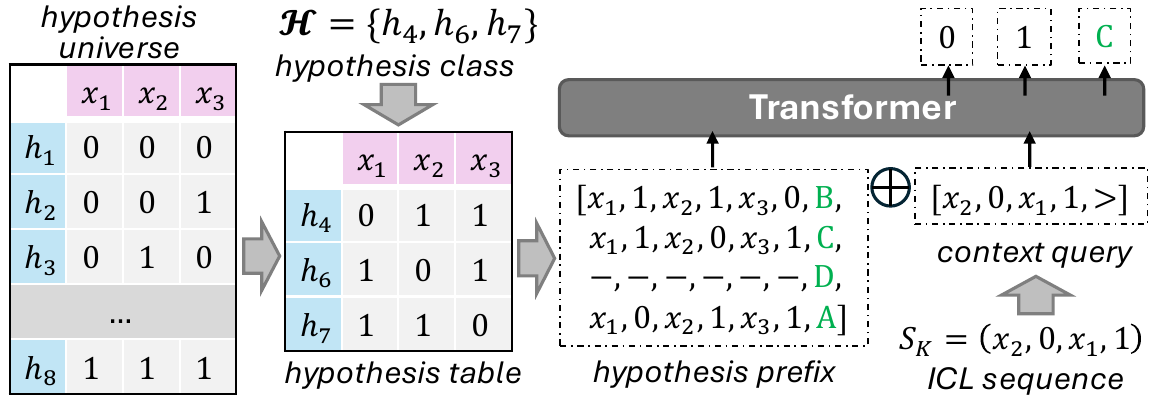}
    \caption{\textbf{Learning ICL-HCG via Transformer.}
    We begin by sampling a subset from the hypothesis universe as the hypothesis class $\mathcal{H}$.
    Next, we encode the hypothesis class $\mathcal{H}$ and concatenate it with context query into a unified sequences of token.
    This sequences is fed into a Transformer model for training with next-token prediction, and testing for evaluating the accuracy on $y$ and hypothesis identification.
    (This figure is an simplified illustration. Please refer to Appendix~\ref{app:prefix} and Fig.~\ref{fig:frameworkfull} for the full details.)}
    \label{fig:framework}
\end{figure}
\subsection{Learning ICL-HCG via Transformer}
\label{subsec:framework}
This section details how Transformer learns ICL-HCG.
As shown in Fig.~\ref{fig:framework},
the hypothesis class $\mathcal{H}$ is first converted to a hypothesis prefix with randomly assigned hypothesis indexes, then concatenated with context query representing sequence $\SK$ as a unified sequence $s$.


\paragraph{Hypothesis prefix\footnote{Please refer to Appendix~\ref{app:prefix} for the full version.}}
Given a hypothesis class $\mathcal{H}=\{h_4,h_6,h_7\}$, its hypothesis prefix with size $L=4$ is constructed as shown in Fig.~\ref{fig:framework}.
Blank hypothesis is used to fill the hypothesis prefix when $|\mathcal{H}|<L$.
A randomly assigned hypothesis index token \hz is used to label each hypothesis.
Leveraging Fig.~\ref{fig:framework} for $L=4$, {\hz}'s are assigned from a pool \{``\textcolor[RGB]{0,176,80}{A}'',``\textcolor[RGB]{0,176,80}{B}'',``\textcolor[RGB]{0,176,80}{C}'',``\textcolor[RGB]{0,176,80}{D}''\} of size $L$ without replacement\footnote{We use variable $z$ to represent the hypothesis index, and create a set of $L$ hypothesis index tokens as a pool from which each hypothesis is randomly assigned a unique index without replacement.}.

\paragraph{Context query}
Given an ICL sequence $\SK$, we append a query token ``\textcolor[RGB]{192,79,21}{\textgreater}'' after it to trigger trigger the prediction of the hypothesis index ss shown in Fig.~\ref{fig:framework}.
We name the combination of $\SK$ and ``\textcolor[RGB]{192,79,21}{\textgreater}'' as context query.

The Transformer predicts the $y$ tokens in the context query based on previous tokens and the index \hz of the underlying hypothesis based on all tokens in the sequence.
The training loss in Eq.~\ref{eq:loss} is further extended to all the tokens in the sequence and implemented as below:
\begin{align}
\label{eq:lossTF}
    \mathcal{L} = - \sum_{t=1}^{T} \log P_\theta(s_i \mid s_{<i}).
\end{align}
We summarize the pipeline in the Appendix~\ref{app:alg} Algorithm~\ref{alg:framework}.




\section{Experiments}
\label{sec:exp}
\begin{figure*}[th!]
    \centering
    \includegraphics[width = 0.85\textwidth]{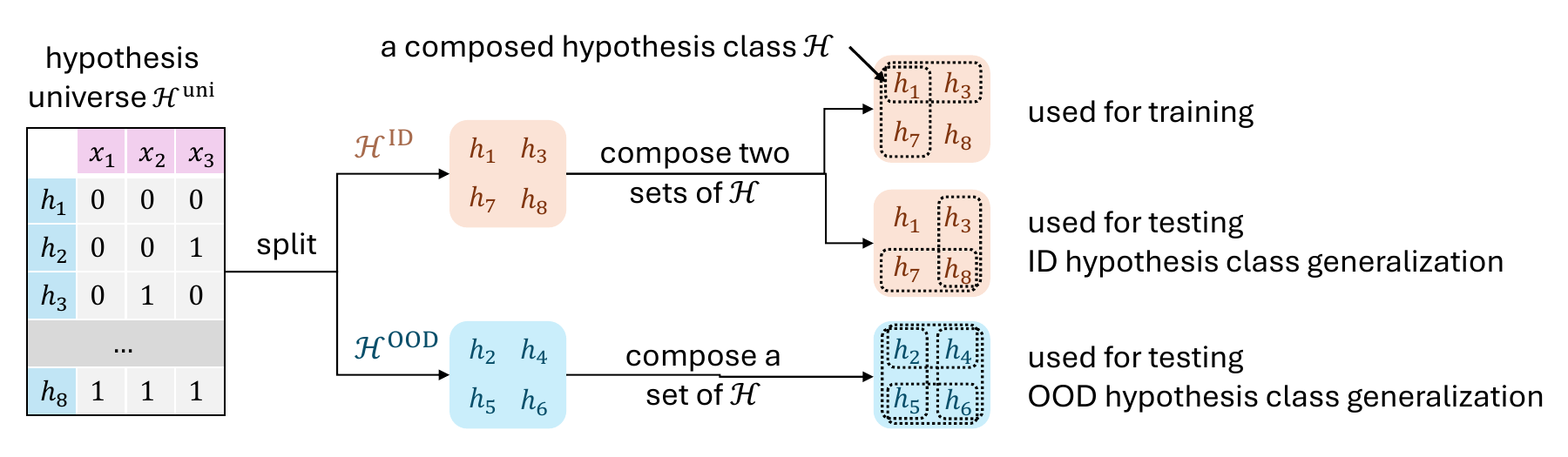}
    \caption{\textbf{The generation of training and testing hypothesis classes.} The hypothesis universe is partitioned into two pools: one for generating training and ID testing hypothesis classes, and another for generating OOD testing hypothesis classes.}
    \label{fig:pipeline}
\end{figure*}
\subsection{Setting of Experiments}
\label{subsec:setting}
\paragraph{Hypothesis class generation} Fig.~\ref{fig:pipeline} illustrates the hypothesis class generation process used in this paper.
We partition the hypothesis universe into two pools: one for generating training and ID testing classes, and another for generating OOD testing hypothesis classes.
This ensures that training and ID testing hypothesis classes do not overlap and that OOD hypothesis classes come from an entirely separate set of hypotheses.
Consequently, both ID and OOD hypothesis class generalization can be assessed using the same trained model.
For detailed realizations of setups for four kinds of generalization, see Appendix~\ref{setup:generalization}.

\paragraph{Pretraining} During pretraining, we backpropagate gradients \emph{based on next-token prediction for all tokens}. 
Each training sequence \(s\) consists of a hypothesis prefix, a context query, and a hypothesis index token. 
As illustrated in Fig.~\ref{fig:framework}, we feed the entire sequence \(s\) (excluding the final index token \hz 
into the Transformer.
We then compute cross-entropy loss for each subsequent token (excluding the very first).
Refer to Appendix~\ref{app:expsetup} for training details, including the learning rate schedule, and hyperparameter search. 

\paragraph{Components of pretraining loss} We conducted experiments to determine the optimal components to include in the pretraining loss. Specifically, we evaluated four configurations: applying the loss (i) solely to the final hypothesis index token, (ii) exclusively to the content query, (iii) only to the label $y$ of the content query, and (iv) across all tokens.
We empirically find that incorporating the loss across all tokens in the sequence leads to the best performance.


\begin{figure}[h!]
\centering
    \subfigure[testing curves of ID hypothesis class generalization.\label{fig:multiple_curves_I_1x1}]
    {
        \includegraphics[width=0.22\textwidth]{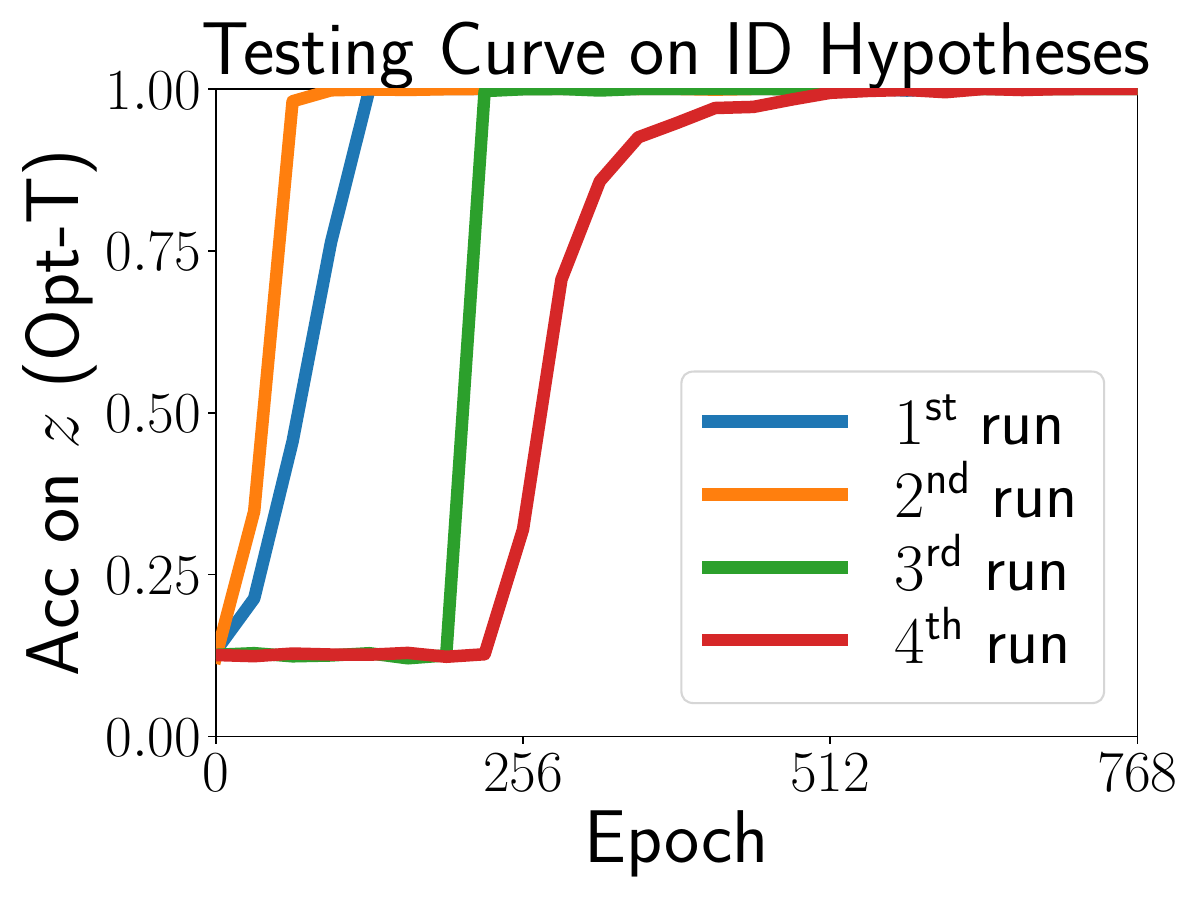}
    }
    \subfigure[testing curves of OOD hypothesis class generalization.\label{fig:multiple_curves_O_1x1}]{
        \includegraphics[width=0.22\textwidth]{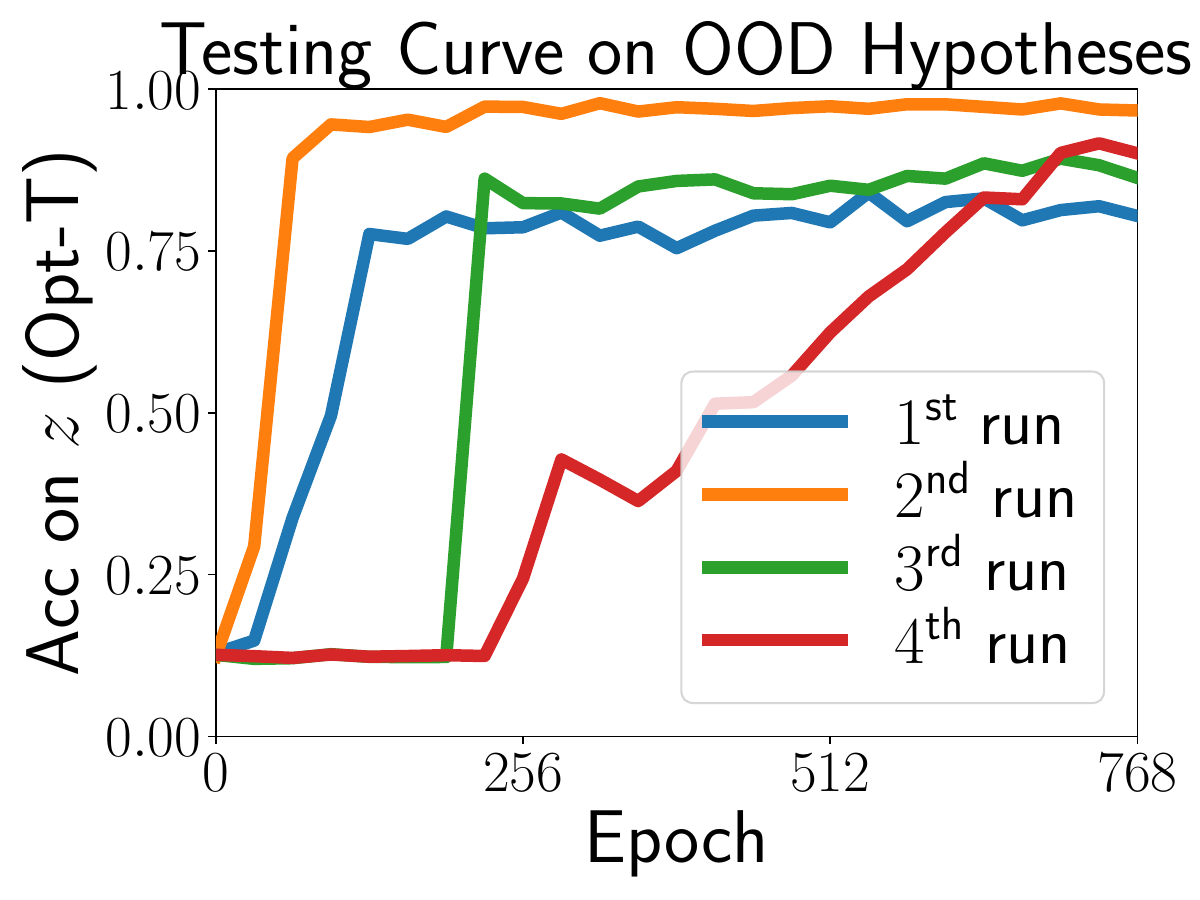}
        }
    \caption{\textbf{Multiple runs on ID and OOD hypothesis class generalizations.} (Different runs imply training and testing with different random seeds.) Transformer successfully learns ICL-HCG, and generalizes to new hypothesis classes and hypotheses.
    Generalization on ID hypotheses is easier than on OOD hypotheses.
    Refer to Appendix~\ref{subapp:4generalization}, Fig.~\ref{fig:multiple_curves_IO_2x3} for more curves of loss, training and testing accuracy.
    }
    \label{fig:multiple_curves_IO}
\end{figure}
\begin{figure}[h!]
\centering
    \subfigure[testing curves of ID hypothesis class size generalization.\label{fig:multiple_curves_IS_1x1}]
    {
        \includegraphics[width=0.475\textwidth]{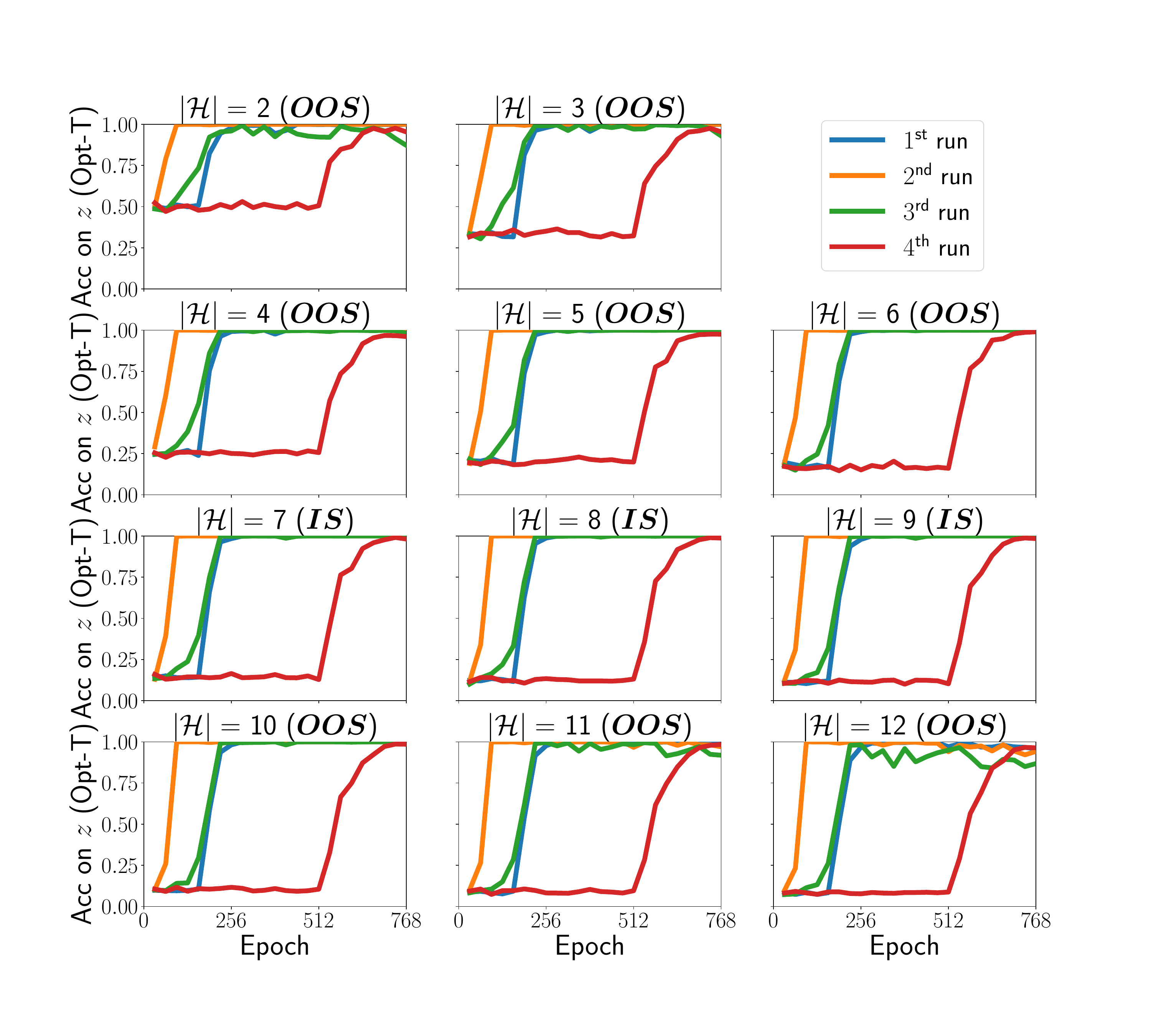}
    }
    \subfigure[testing curves of OOD hypothesis class size generalization.\label{fig:multiple_curves_OS_1x1}]{
        \includegraphics[width=0.475\textwidth]{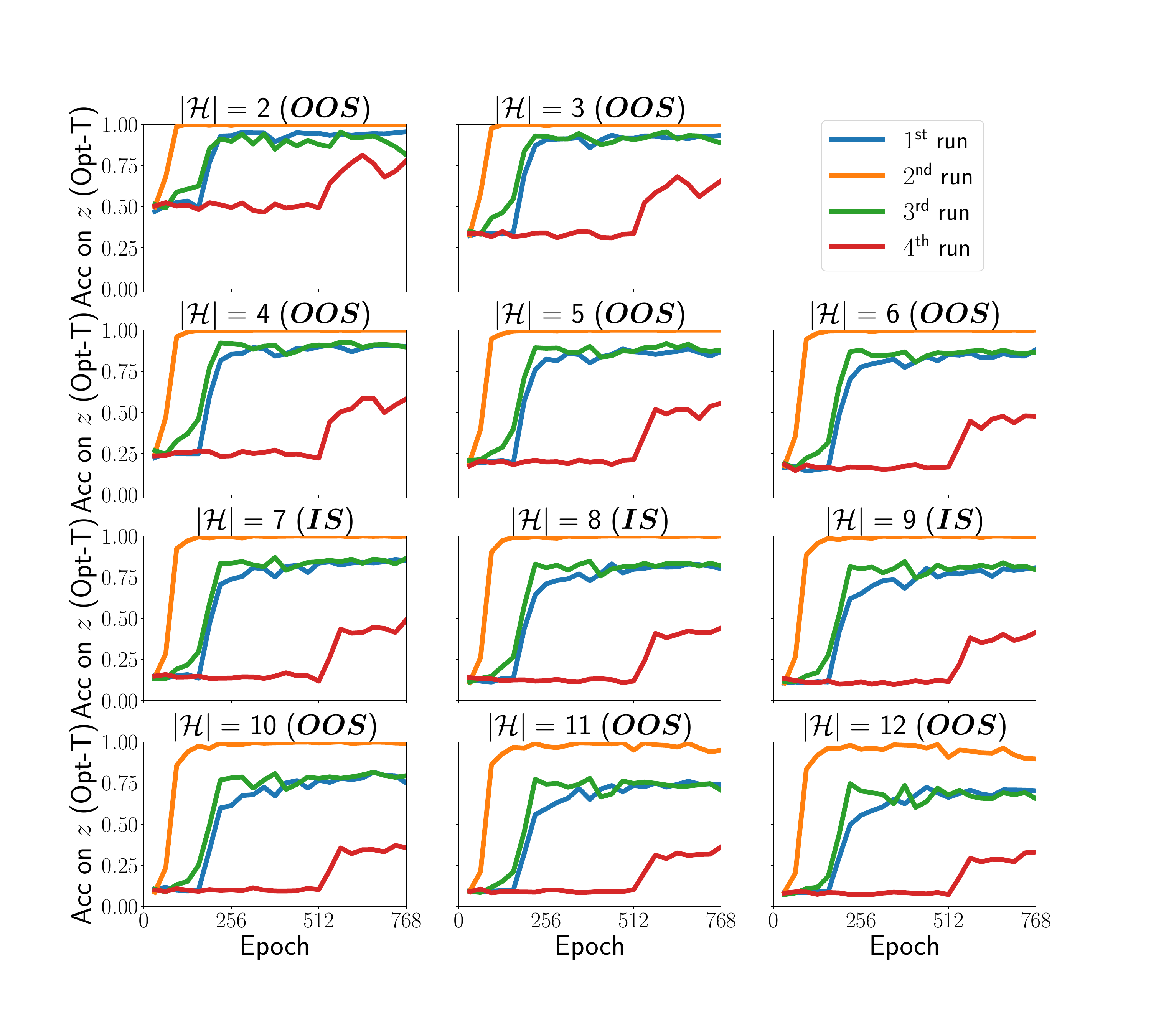}
        }
    \caption{\textbf{Multiple runs on ID and OOD hypothesis class size generalizations.}
    (Different runs imply training and testing with different random seeds.)
    Transformers trained on hypothesis classes with sizes \(\lvert \mathcal{H} \rvert \in \{7,8,9\}\) successfully generalizes to hypothesis classes with sizes \(\lvert \mathcal{H} \rvert \in \{2,3,\ldots,13,14\}\) under ID hypothesis class size generalization.
    In contrast, the same trained Transformer exhibits poorer performance on OOD hypothesis class size generalization.
    In the figure, \emph{IS} stands for ``in-size,'' indicating the hypothesis class sizes included in the training, while \emph{OOS} stands for ``out-of-size,'' indicating the sizes that are \textbf{not} included in the training.
    Refer to Appendix~\ref{subapp:4generalization}, Fig.~\ref{fig:multiple_curves_IOS_9x5} for training accuracy curves.
    }
    \label{fig:multiple_curves_IOS}
\end{figure}
\subsection{Four Types of Generalization}
\label{subsec:4generalization}
This section investigates whether a Transformer trained on ICL-HCG tasks can generalize to new tasks, \ie, new hypothesis classes.
We explore four types of generalization scenarios, defined in Definitions~\ref{def:SpaceGeneralization},~\ref{def:HypothesisGeneralization},~\ref{def:Space&SizeGeneralization}, and~\ref{def:Hypothesis&SizeGeneralization}. Detailed hyperparameters of settings are provided in Appendix~\ref{setup:generalization}.
\begin{highlight}
    \paragraph{Finding 1:} 
    \emph{Transformer can successfully learn ICL-HCG tasks and such a learned ability can generalize to new hypothesis, hypothesis class, and hypothesis size, whereas the generalization on OOD hypotheses is harder than ID hypotheses.}
\end{highlight}
We first demonstrate that the Transformer successfully learns ICL-HCG and that this capability generalizes effectively on ID and OOD hypothesis class generalizations.
As illustrated in Figs.~\ref{fig:multiple_curves_I_1x1} and~\ref{fig:multiple_curves_O_1x1}, the Trained Transformers on 4 runs with different random seeds all achieve near-perfect accuracy (close to 1.00) on ID hypothesis class generalization, and around 0.8 to 0.9 accuracy on OOD hypothesis class generalization.
Furthermore, we show that the learned ICL-HCG ability generalizes to hypothesis classes of various sizes.
As depicted in Figs.~\ref{fig:multiple_curves_IS_1x1} and~\ref{fig:multiple_curves_OS_1x1}, the trained Transformers achieve near 1.00 accuracy for $|\mathcal{H}|\in\{2,\ldots,12\}$ on ID hypothesis class size generalization, while exhibiting moderately lower accuracy on OOD hypothesis class size generalization.
Both Figs.~\ref{fig:multiple_curves_IO} and~\ref{fig:multiple_curves_IOS} indicate that generalization on OOD hypotheses is more challenging compared to ID hypotheses.
\subsection{Model Architecture Comparisons}
\label{subsec:4model}
\begin{figure}[h!]
\centering
    \subfigure[testing curves of ID hypothesis class size generalization.\label{fig:multiple_models_IS_4x3}]
    {
        \includegraphics[width=0.5\textwidth]{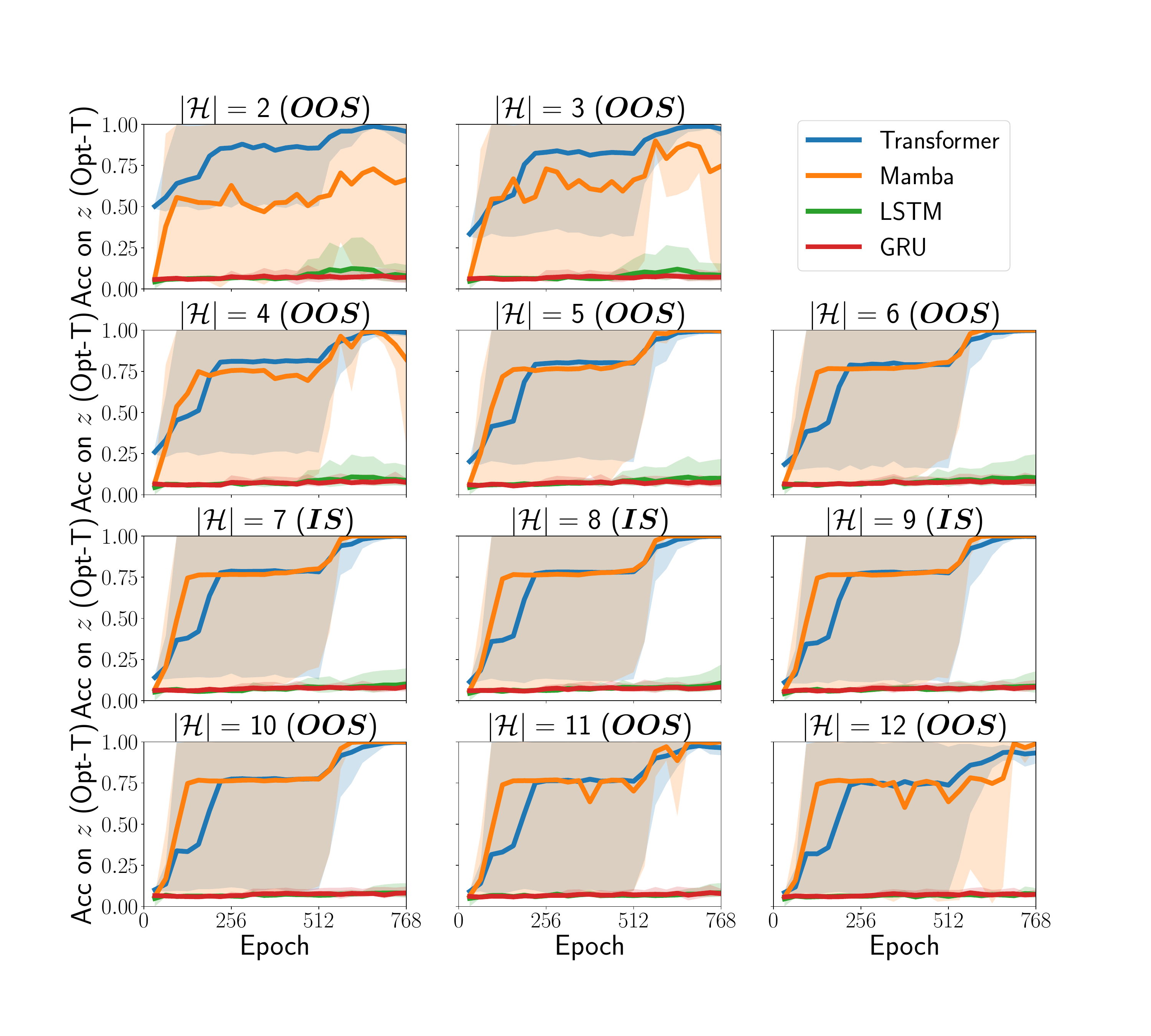}
    }
    \hspace{0.1em}
    \subfigure[testing curves of OOD hypothesis class size generalization.\label{fig:multiple_models_OS_4x3}]{
        \includegraphics[width=0.5\textwidth]{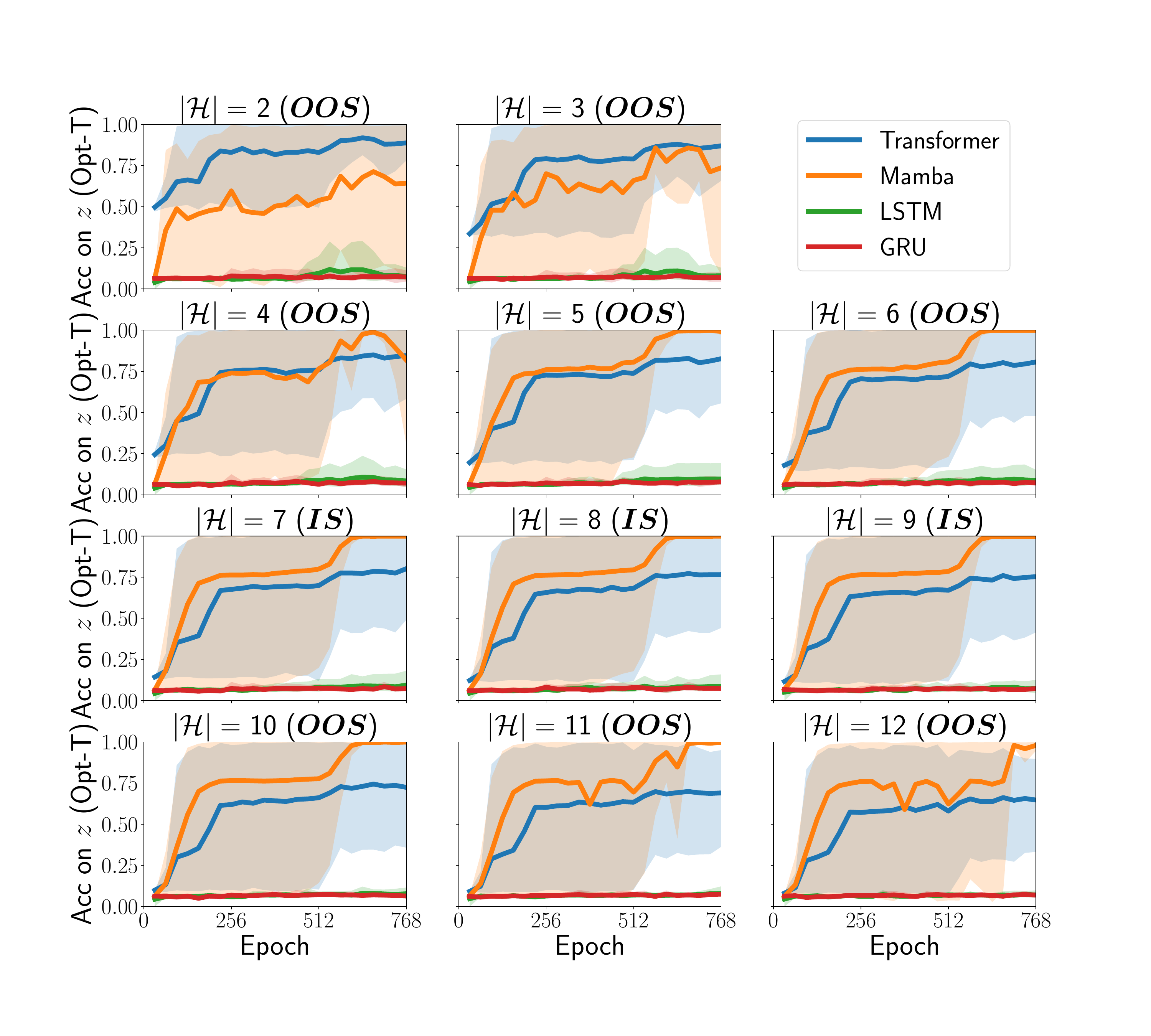}
        }
    \caption{
    \textbf{Various models on ID and OOD hypothesis class size generalizations.}
    In both settings, Transformers and Mamba exhibit strong generalization, whereas LSTM and GRU fail to do so. For hypothesis class sizes \(\lvert \mathcal{H} \rvert \in \{7,8,9\}\), Mamba achieves accuracy comparable to Transformer on ID hypothesis class generalization, and surpasses Transformer on OOD hypothesis class generalization.
    However, Transformers show similar or higher accuracy than Mamba on ID hypothesis class size generalization, suggesting a potential advantage in length generalization.
    Refer to Appendix~\ref{subapp:4model}, Fig.~\ref{fig:multiple_models_IOS_9x5} for training accuracy curves.
    }
    \label{fig:multiple_models_IOS}
\end{figure}

We compare Transformer with other model architectures, including Mamba~\cite{gu2023mamba}, LSTM~\cite{hochreiter1997long}, and GRU~\cite{cho2014learning}. We investigate whether each model can effectively fit the training dataset and, if so, generalize to the four types of unseen hypothesis classes.
\begin{figure}[h!]
\centering
    \subfigure[testing curves of ID hypothesis class generalization.\label{fig:multiple_models_I_1x1}]
    {
        \includegraphics[width=0.22\textwidth]{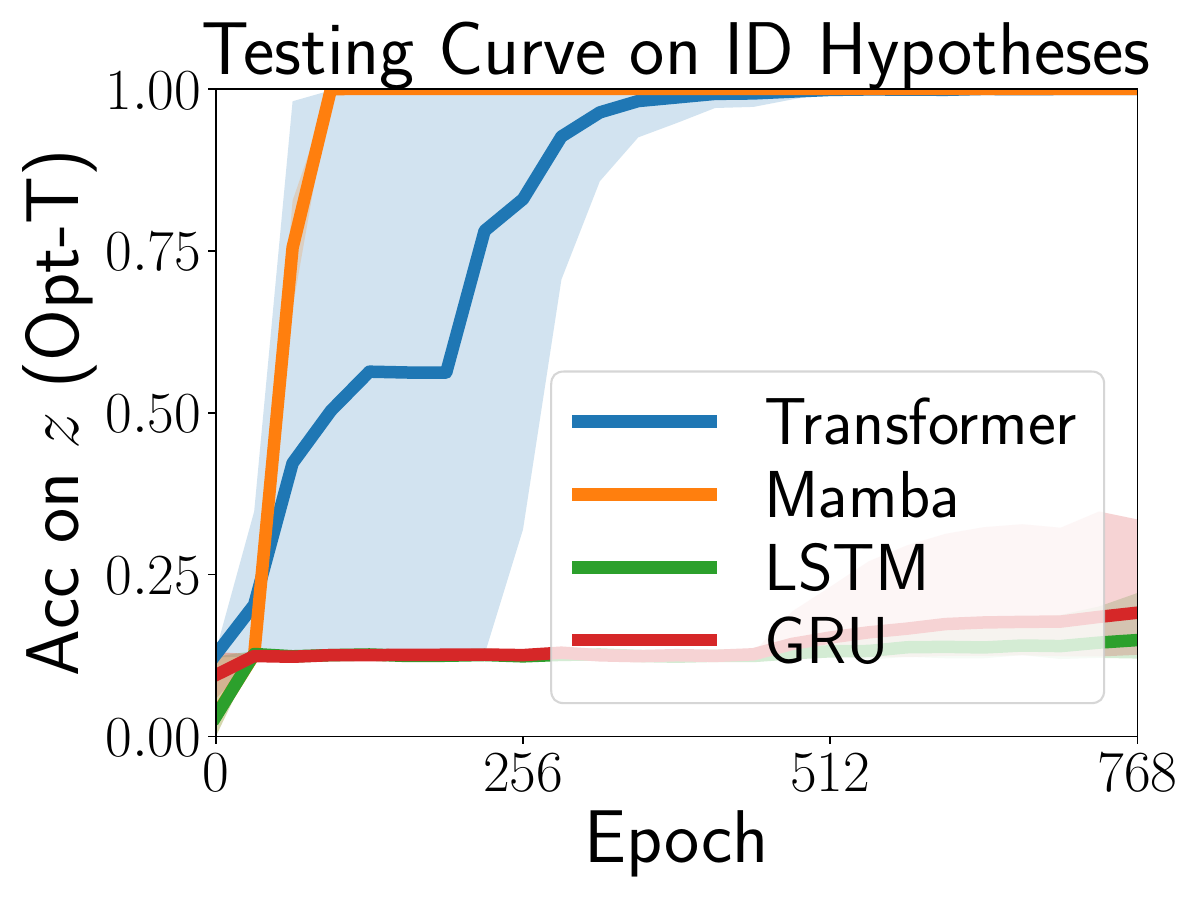}
    }
    \hspace{0.1em}
    \subfigure[testing curves of OOD hypothesis class generalization.\label{fig:multiple_models_O_1x1}]{
        \includegraphics[width=0.22\textwidth]{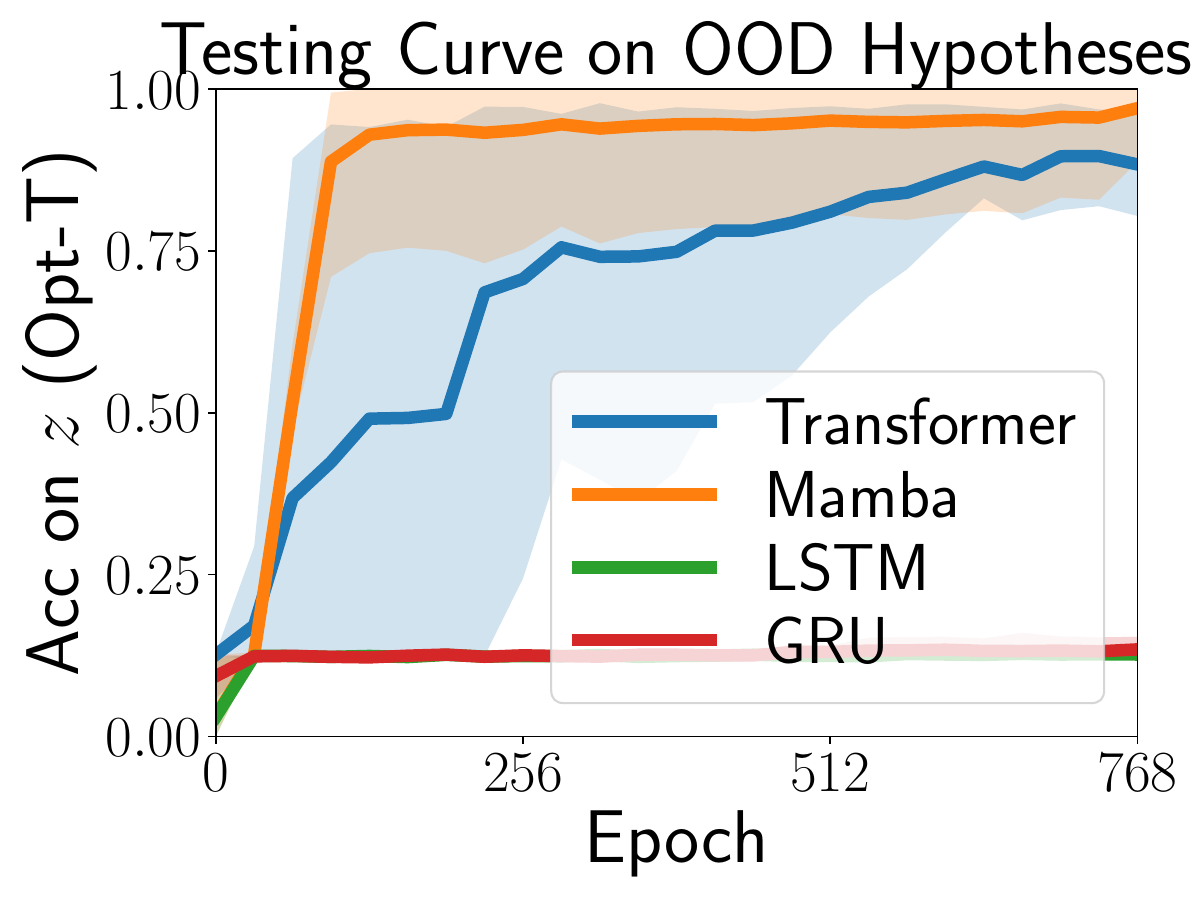}
        }
    \caption{\textbf{Various models on ID and OOD hypothesis class generalizations.}
    Transformer and Mamba succeed on both ID and OOD hypothesis class generalization, whereas LSTM and GRU fail.
    Mamba exhibits slightly higher accuracy than Transformer on OOD generalization. Refer to Appendix~\ref{subapp:4model} and Fig.~\ref{fig:multiple_models_IO_2x3} for training curves.}
    \label{fig:multiple_models_IO}
\end{figure}
\begin{highlight}
    \paragraph{Finding 2:} 
    \emph{While both Mamba and Transformer excel on the four generalization tasks, LSTM and GRU fail to handle the ICL-HCG tasks. Mamba outperforms Transformer on OOD hypothesis class generalization, whereas Transformer outperforms Mamba on ID hypothesis class size generalization.}
\end{highlight}
We evaluate ID and OOD hypothesis class generalization across model architectures. 
Within the hyperparameter search space in Appendix~\ref{app:hyperparameters}, Fig.~\ref{fig:multiple_models_IO} shows that Transformer and Mamba both generalize well on ID and OOD hypothesis class generalizations, with higher accuracy on ID hypotheses (1.00 accuracy) than OOD (around 0.8 to 0.9 accuracy).
In contrast, LSTM and GRU fail to fit the task, achieving approximately 0.125 accuracy, equivalent to random guessing over eight hypotheses.
Furthermore, Fig.~\ref{fig:multiple_models_IOS} shows that Mamba outperforms Transformer on OOD hypothesis class size generalization, whereas Transformer excels on ID hypothesis class size generalization, suggesting a potential advantage of Transformer on length generalization, and Mamba on generalization of OOD hypotheses.

\subsection{Effect of Training Hypothesis Class Count}
\label{subsec:numtrain}
\begin{figure}[h!]
\centering
    \subfigure[testing curves of ID hypothesis class generalization.\label{fig:num_train_I_1x1}]
    {
        \includegraphics[width=0.225\textwidth]{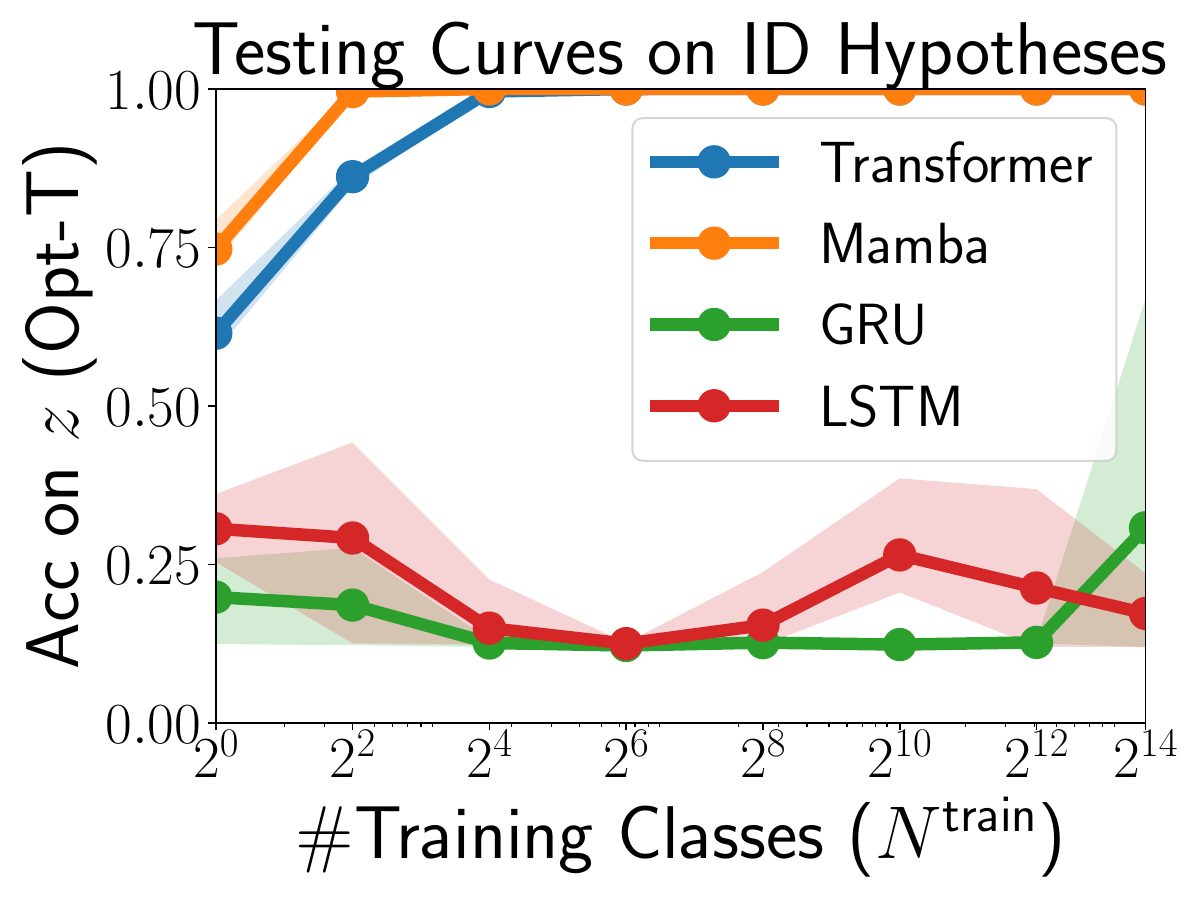}
    }
    \subfigure[testing curves of OOD hypothesis class generalization.\label{fig:num_train_O_1x1}]{
        \includegraphics[width=0.225\textwidth]{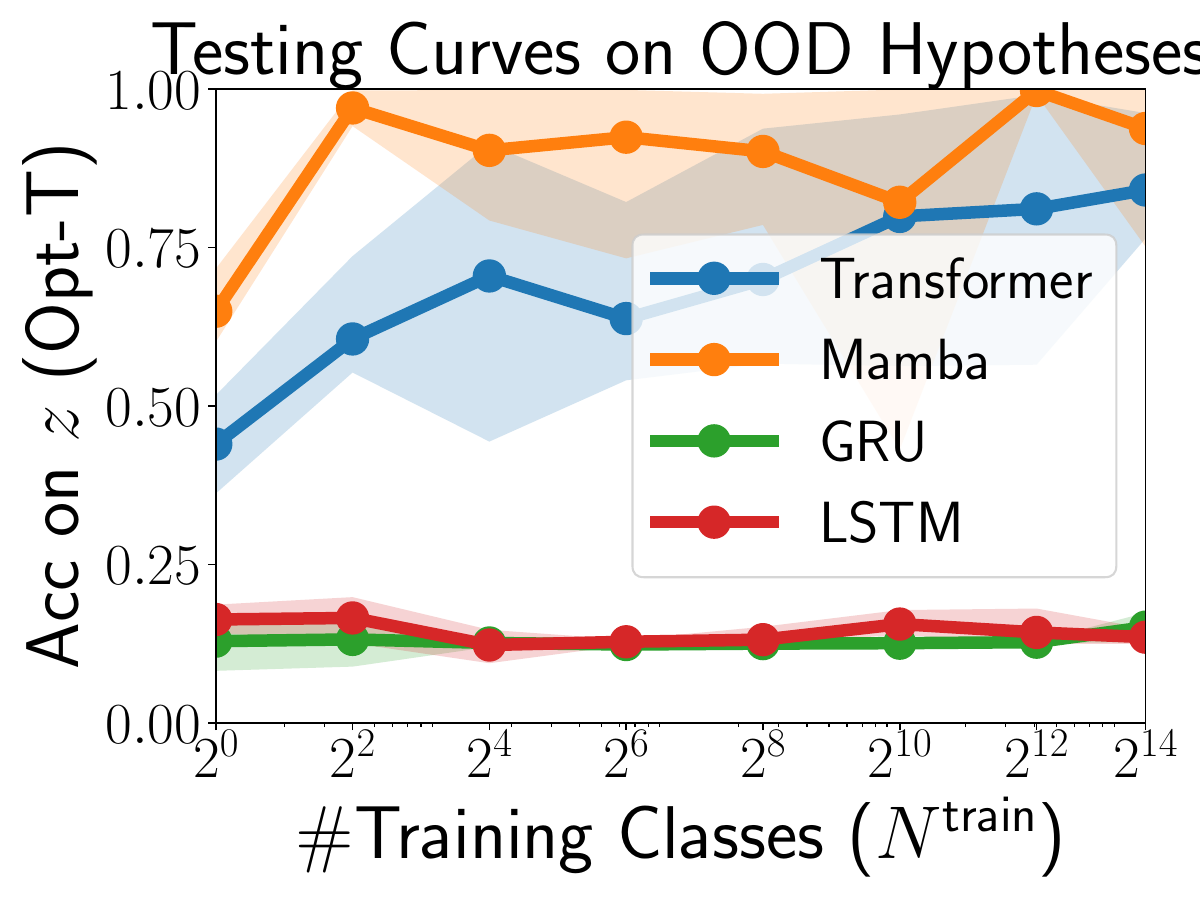}
        }
    \caption{\textbf{Effect of training hypothesis class count.}
    Transformer and Mamba trained on ICL-HCG tasks generalize to new hypothesis classes with only $4$ to $16$ training hypothesis classes. 
    Refer to Appendix~\ref{subapp:numtrain}, Fig.~\ref{fig:num_train_IO_1x5} for training accuracy and more details.}
    \label{fig:num_train_IO}
\end{figure}
We evaluate how the number of training hypothesis classes affects ID and OOD hypothesis class generalization abilities.
\begin{highlight}
    \paragraph{Finding 3:} 
    \emph{Mamba is more sample efficient than Transformer on ICL-HCG tasks, and achieves near perfect generalization with only several pretraining hypothesis classes.}
\end{highlight}
In Fig.~\ref{fig:num_train_I_1x1}, we evaluate Mamba, Transformer, GRU, and LSTM. 
With only \(2^2\) and \(2^4\) training hypothesis classes, Mamba and Transformer 
achieve near-perfect (1.00 accuracy) ID hypothesis class generalization, while LSTM and GRU fail 
to fit the ICL-HCG tasks.
In Fig.~\ref{fig:num_train_O_1x1}, Mamba nearly achieves 
perfect OOD hypothesis class generalization with as few as \(2^2\) training classes, 
whereas Transformer's accuracy improves gradually with more training classes.

\subsection{Effect of Imbalanced In-Context Samples}
\label{subsec:numtrain}
\begin{figure}[h!]
    \centering
    \includegraphics[width = 0.475\textwidth]{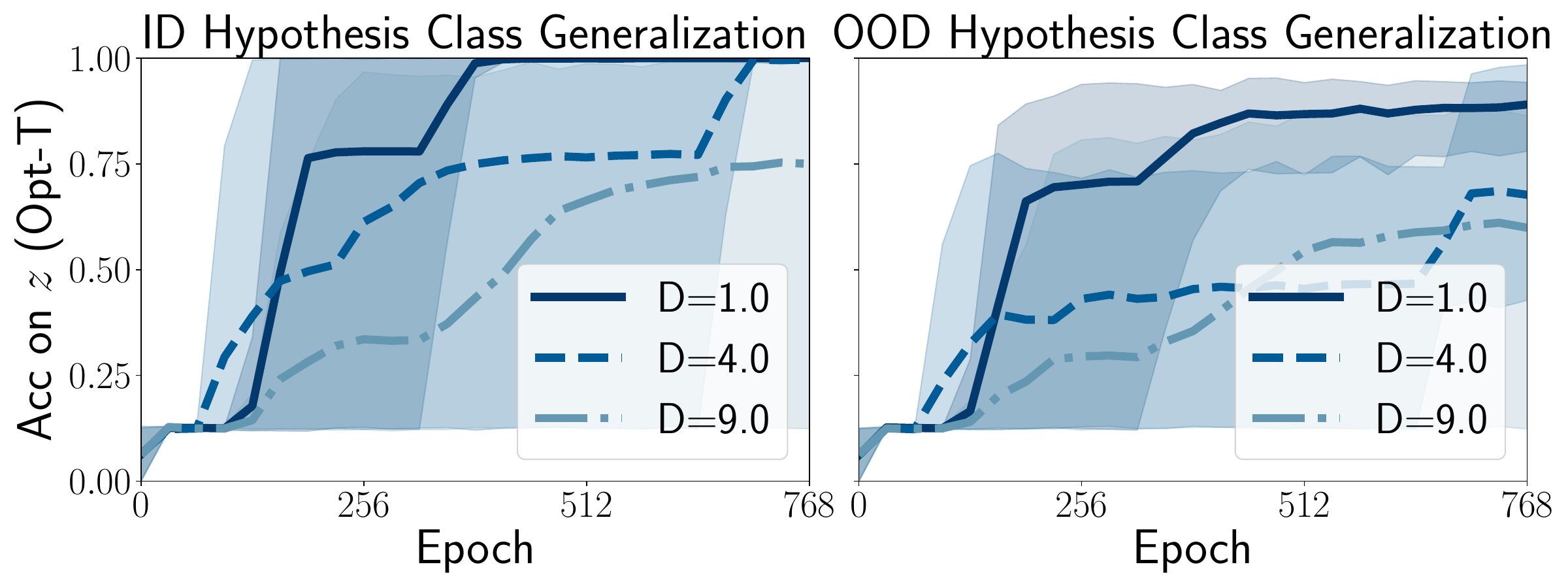}
    \caption{\textbf{The effect of sample imbalance.}
    Sample imbalance leads to lower convergence speed.}
    \label{fig:Imbalance}
\end{figure}

This section investigates how an imbalanced sample distribution in the context query affects the training procedure.
Specifically, we consider the following distribution over $\mathcal{X}$:
$$
\text{norm}\left(\frac{1}{\sqrt{\text{D}}},\ldots,\frac{1}{\sqrt{\text{D}}},1,\sqrt{\text{D}},\ldots,\sqrt{\text{D}}\right),
$$
where the first half of the terms are $\frac{1}{\sqrt{\text{D}}}$, the middle term (if $|\mathcal{X}|$ is odd) is $1$, the second half consists of $\sqrt{\text{D}}$, and
D\footnote{The notation D is distinguished from token ``\textcolor[RGB]{0,176,80}{D}'' by color and dataset $\mathcal{D}$ by format.} represents the disparity of the distribution over $\mathcal{X}$, \ie, $\text{D}=\frac{\max_{x\in\mathcal{X}} P(x)}{\min_{x\in\mathcal{X}} P(x)}$.
\begin{highlight}
    \paragraph{Finding 4:} 
    \emph{In-context sample imbalance lags the convergence of training.}
\end{highlight}
We analyzed the impact of imbalance on the training process in Fig.~\ref{fig:Imbalance} by varying D values, showing that greater imbalance slows convergence.
On average over four runs, training converges in about 384 epochs for $D=1$ but takes around 700 epochs for D$=4$.
\subsection{The Benefit of Hypothesis Prefix}
\label{subsec:icl}
\begin{figure}[h!]
    \centering
    \includegraphics[width = 0.475\textwidth]{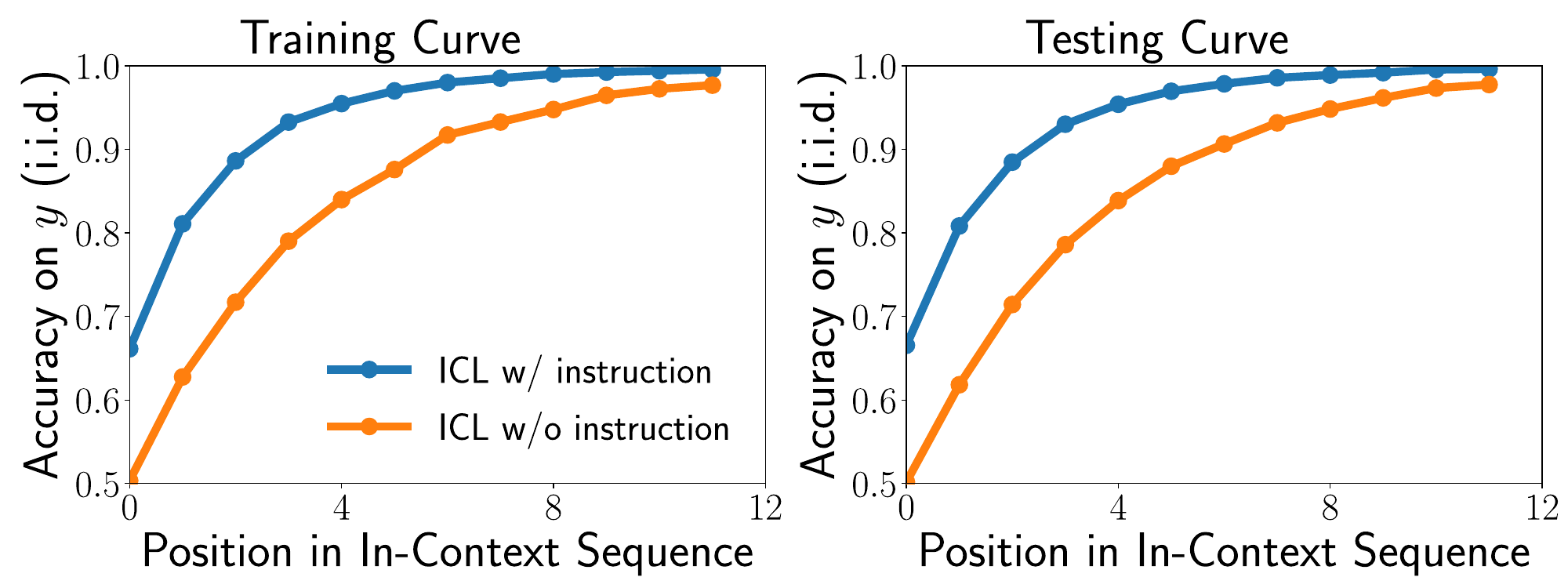}
    \caption{\textbf{The effect of instruction.}
    Under ID hypothesis class generalization, providing an instruction (hypothesis prefix) significantly boosts ICL performance, especially when the $y$ token appears early (indicating only a few demonstration examples precede it).
    }
    \label{fig:ICL}
\end{figure}

In this section, we demonstrate how the hypothesis prefix influences the accuracy of ICL.
We compare ICL accuracy on $y$ with hypothesis prefix and without hypothesis prefix, under the setting of ID hypothesis class generalization.
\begin{highlight}
    \paragraph{Finding 5:} 
    \emph{Incorporating hypothesis prefix as instruction significantly boost the accuracy of ICL.}
\end{highlight}
As shown in Fig.~\ref{fig:ICL}, the hypothesis prefix significantly enhances the training and testing accuracy on $y$ of ICL.
Using position 3 as an example, predictions with three $(x,y)$ pairs as demonstrations achieve approximately 0.95 accuracy with instruction but only around 0.8 without, highlighting the effectiveness of instruction.
\subsection{The Effect of Pretraining Hypothesis Diversity}
\label{subsec:diversity}
\begin{figure}[h!]
    \centering
    \includegraphics[width = 0.475\textwidth]{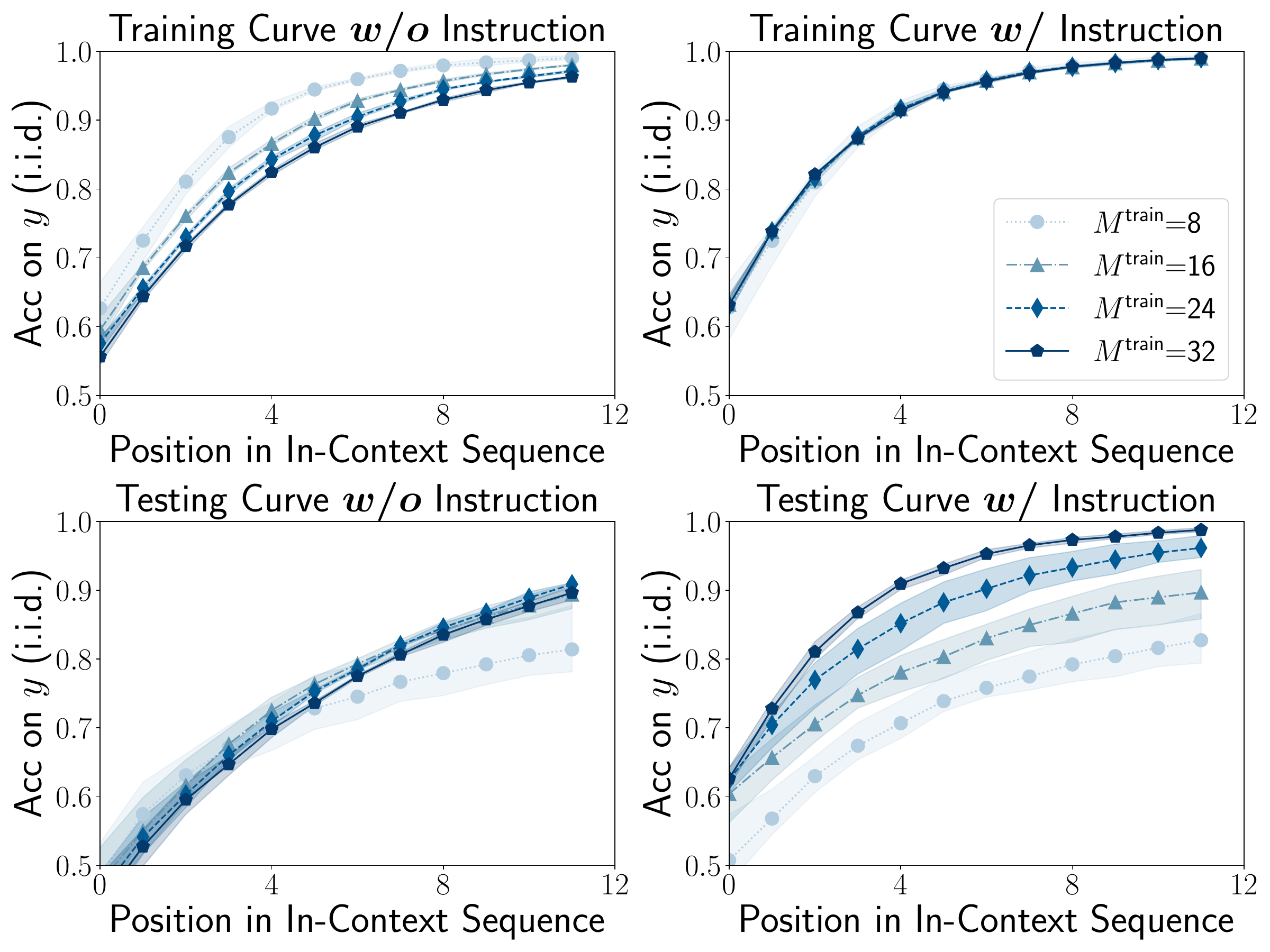}
    \caption{
    \textbf{The effect of pretraining hypothesis diversity.} Under hypothesis generalization, increasing the diversity of pretraining hypotheses significantly boosts the performance of ICL when instructions are provided. However, without instructions, this effect is limited.
    }
    \label{fig:diversity}
\end{figure}

In this section, we investigate the impact of hypothesis diversity combined with instruction (hypothesis prefix) on ICL accuracy.  
We conduct experiments under OOD hypothesis class generalization with an input space size of $|\mathcal{X}| = 6$, leading to a hypothesis universe $\mathcal{H}^{\text{uni}}$ of $2^{|\mathcal{X}|} = 64$ hypotheses.  
$\mathcal{H}^{\text{uni}}$ is split into $\mathcal{H}^{\text{ID}}$ with 48 hypotheses and $\mathcal{H}^{\text{OOD}}$ with 16.  
For training, we sample $M^{\text{train}} \in \{8, 16, 24, 32\}$ hypotheses from $\mathcal{H}^{\text{ID}}$ to examine the effect of training hypothesis diversity, while testing uses all hypotheses in $\mathcal{H}^{\text{ID}}$.
\begin{highlight}
    \paragraph{Finding 6:} 
    \emph{Increasing the diversity of pretraining hypotheses significantly boosts the performance of ICL when instructions are provided.}
\end{highlight}
As illustrated in Fig.~\ref{fig:diversity}, under OOD hypothesis class generalization, the Transformer trained with instructions achieves similar ICL accuracy to a standard ICL approach when pretraining hypothesis diversity is low, but significantly outperforms it when pretraining hypothesis diversity is high.
Using position 10 as an example, with instruction, increasing $M^{\text{train}}$ from 8 to 32 raises accuracy from 0.80 to approximately 0.99.  
Without instruction, the same increase in diversity improves accuracy only from 0.80 to 0.90.
Notably, the testing ICL samples are derived from unseen hypotheses, indicating that incorporating instructions can enhance ICL performance for new hypotheses.

\section{Discussion}  
Building on the ICL-HCG framework, we conduct diverse experiments focusing on generalization.
While Bayesian inference~\cite{xie2021explanation} offers insights into ICL, prior work~\citep{raventos2024pretraining} has shown that Transformers can generalize beyond Bayesian inference with sufficient pretraining task diversity.
However, the mechanisms underlying such OOD generalization remain unclear. Our work provides a framework for exploring OOD generalization beyond Bayesian inference, where no test samples appear in the training set due to disjoint hypothesis classes.

Furthermore, in Sec.~\ref{subsec:4model}, we demonstrate that Transformer and Mamba exhibit distinct strengths: Transformer excels on length generalization, while Mamba performs better on OOD hypothesis generalization.  
In Sec.~\ref{subsec:diversity}, we show that instruction enhances the benefits of pretraining hypothesis diversity.  
These findings highlight two key factors influencing OOD generalization: (i) model architecture and (ii) data structure.
Future work will further explore these phenomena, focusing on understanding the underlying mechanisms of OOD generalization in Transformer and Mamba.

\section{Conclusion}
In this paper, we introduced a novel instruction-based ICL framework that explicitly integrates a hypothesis class as the instruction, namely ICL-HCG.
Through a series of diverse experiments, we demonstrated that Transformers trained on ICL-HCG tasks can generalize to new hypothesis classes and new hypotheses, even when trained on only a few hypothesis classes.
Moreover, we show that the incorporation of such instructions significantly enhances the accuracy of ICL, thereby bridging the gap between synthetic ICL studies and real-world applications.
We also examined the effect of hypothesis diversity within this framework and found that increased hypothesis diversity substantially improves ICL accuracy especially when instructions are provided.
Our framework serves as a platform for diverse explorations, offering new insights and serving as a playground for research on ICL and LLMs.

We conclude our paper by acknowledging the limitations of our current framework: (i) Our study is confined to finite hypothesis binary classification problems, which can be extended to more complex scenarios;
(ii) The hypothesis prefix is assumed to provide an explicit hypothesis class, differing from the more implicit instructions used in real-world LLM applications.

\section*{Acknowledgements}
This work of Kangwook Lee is supported 
by NSF Award DMS-2023239, NSF CAREER Award CCF-2339978, Amazon Research Award, and a grant from FuriosaAI.



\bibliography{example_paper}
\bibliographystyle{icml2025}

\appendix
\onecolumn
\section{Pseudo Algorithm for ICL-HCG}
\label{app:alg}
We summarize our meta framework for ICL-HCG in Algorithm~\ref{alg:framework}.
\begin{algorithm}
  \caption{Meta-Learning Framework for ICL-HCG}
  \label{alg:framework}
  \begin{algorithmic}[1]
    \STATE \textbf{Inputs:} a set of inputs $\mathcal{X}$, a training set of hypothesis classes $\mathcal{S}^{\text{train}}=\{\mathcal{H}_i^{\text{train}}\}_{i=1}^{N^{\text{train}}}$, a testing set of hypothesis classes $\mathcal{S}^{\text{test}}=\{\mathcal{H}_i^{\text{test}}\}_{i=1}^{N^{\text{test}}}$, batch size $B$, hypothesis prefix size $L$, and context query size $K$
    \FOR{\textbf{training epoch}}
      \STATE sample $\{\mathcal{H}_i\}_{i=1}^B \overset{\text{i.i.d.}}{\sim} \text{Uniform}(\mathcal{S}^{\text{train}})$
      \FOR{each hypothesis class $\mathcal{H} \in \{\mathcal{H}_i\}_{i=1}^B$}
        \STATE generate $h,\SK$ following \hyperref[asu:iid]{i.i.d. Generation}
        
        \STATE \textbf{// Construct sequence based on $\mathcal{H}$, $h$, and $\SK$}
        \STATE construct hypothesis prefix, context query, and hypothesis index $z$ based on $\mathcal{H}$, $h$, $\SK$
        \STATE $s \gets \text{concatenate}(\text{hypothesis prefix}, \text{context query}, z)$
    
        \STATE \textbf{// Cross-entropy loss for next token prediction}
        \STATE $\mathcal{L} \gets -\sum_{t=2}^{|s|} \log P(s_t \mid s_{<t})$
      \ENDFOR
      \STATE update model parameters using $\mathcal{L}$ of the batch
    \ENDFOR
    
    \FOR{\textbf{testing epoch}}
      \STATE sample $\{\mathcal{H}_i\}_{i=1}^B \overset{\text{i.i.d.}}{\sim} \text{Uniform}(\mathcal{S}^{\text{test}})$
      
      \FOR{each hypothesis class $\mathcal{H} \in \{\mathcal{H}_i\}_{i=1}^B$}
        \STATE generate $h,\SK$ via:
        \STATE \quad \textbf{either} following \hyperref[asu:iid]{i.i.d. Generation}
        \STATE \quad \textbf{or} following \hyperref[asu:optt]{Opt-T Generation}
        \STATE \textbf{construct sequence $s$ based on $\mathcal{H}$, $h$, and $\SK$}
        \STATE \textbf{evaluate the prediction accuracy on $y$, $z$, etc}
      \ENDFOR
    \ENDFOR
  \end{algorithmic}
\end{algorithm}

\section{Implementation Detail of Hypothesis Prefix and Context Query}
\label{app:prefix}
\begin{figure}[h!]
    \centering
    \includegraphics[width = 0.75\textwidth]{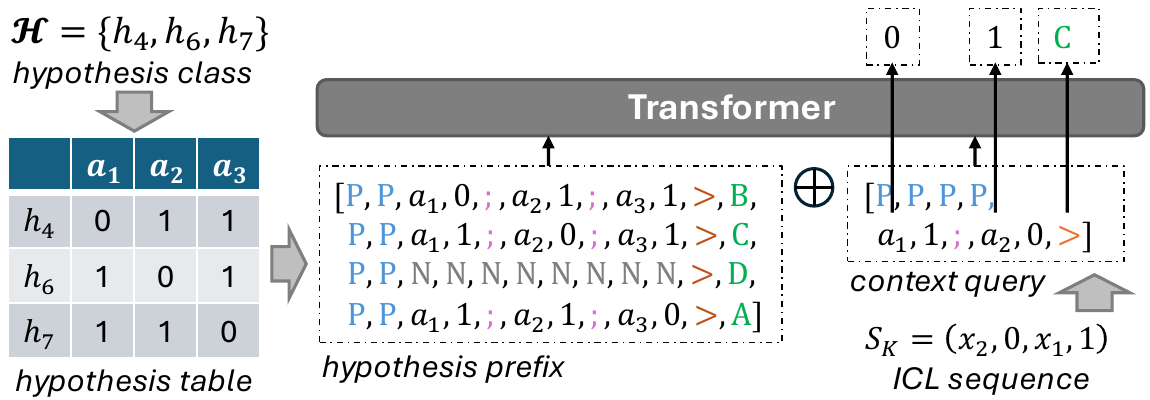}
    \caption{\textbf{The framework.} We convert hypothesis class $\mathcal{H}$ and ICL sequence $\SK$ into sequences of tokens, concatenate them and input to Transformer.
    Then we examine whether Transformer can predict correct $y$ and $z$ values.}
    \label{fig:frameworkfull}
\end{figure}
\paragraph{Hypothesis prefix}
\label{def:HypothesisPrefixFull}
Given a hypothesis class $\mathcal{H}$ and its hypothesis table, the correspongding hypothesis prefix with hypothesis prefix's content length $L$ is constructed as shown in Fig.~\ref{fig:frameworkfull}.
The token ``\textcolor[RGB]{78,149,217}{P}'' serves as the padding token to separate hypotheses,
the token ``\textcolor[RGB]{216,110,204}{;}'' serves as the separation token to separate $(x,y)$ pairs,
the token ``\textcolor[RGB]{127,127,127}{N}'' serves as the empty token to fill a blank hypothesis,
and the token ``\textcolor[RGB]{192,79,21}{\textgreater}'' is used to connect $(x,y)$ pairs of the hypothesis to a randomly assigned hypothesis index \hz\footnote{We use variable $z$ to represent the hypothesis index.}.
In the illustrated example in Fig.~\ref{fig:frameworkfull}, the randomly assigned indexes {\hz}'s are sampled from $M=4$ hypothesis index tokens \{``\textcolor[RGB]{0,176,80}{A}'',``\textcolor[RGB]{0,176,80}{B}'',``\textcolor[RGB]{0,176,80}{C}'',``\textcolor[RGB]{0,176,80}{D}''\} without replacement\footnote{A set of $L$ hypothesis index tokens are created serve as the pool from which the hypothesis indexes are randomly sampled without replacement.}.

\paragraph{Context query}
Given an ICL sequence $\SK$ with $K$ pairs of $(x,y)$, the context query of size $K$ is constructed to represent the ICL sequence and trigger the prediction of the hypothesis index with padding token ``\textcolor[RGB]{78,149,217}{P}'', separation token token ``\textcolor[RGB]{216,110,204}{;}'', and query token ``\textcolor[RGB]{192,79,21}{\textgreater}'' as shown in Fig.~\ref{fig:frameworkfull}.
\section{Additional Details of Experiments}
\label{app:exp}

\subsection{Four Types of Generalization}
\begin{figure*}[th!]
    \centering
    \includegraphics[width = 0.8\textwidth]{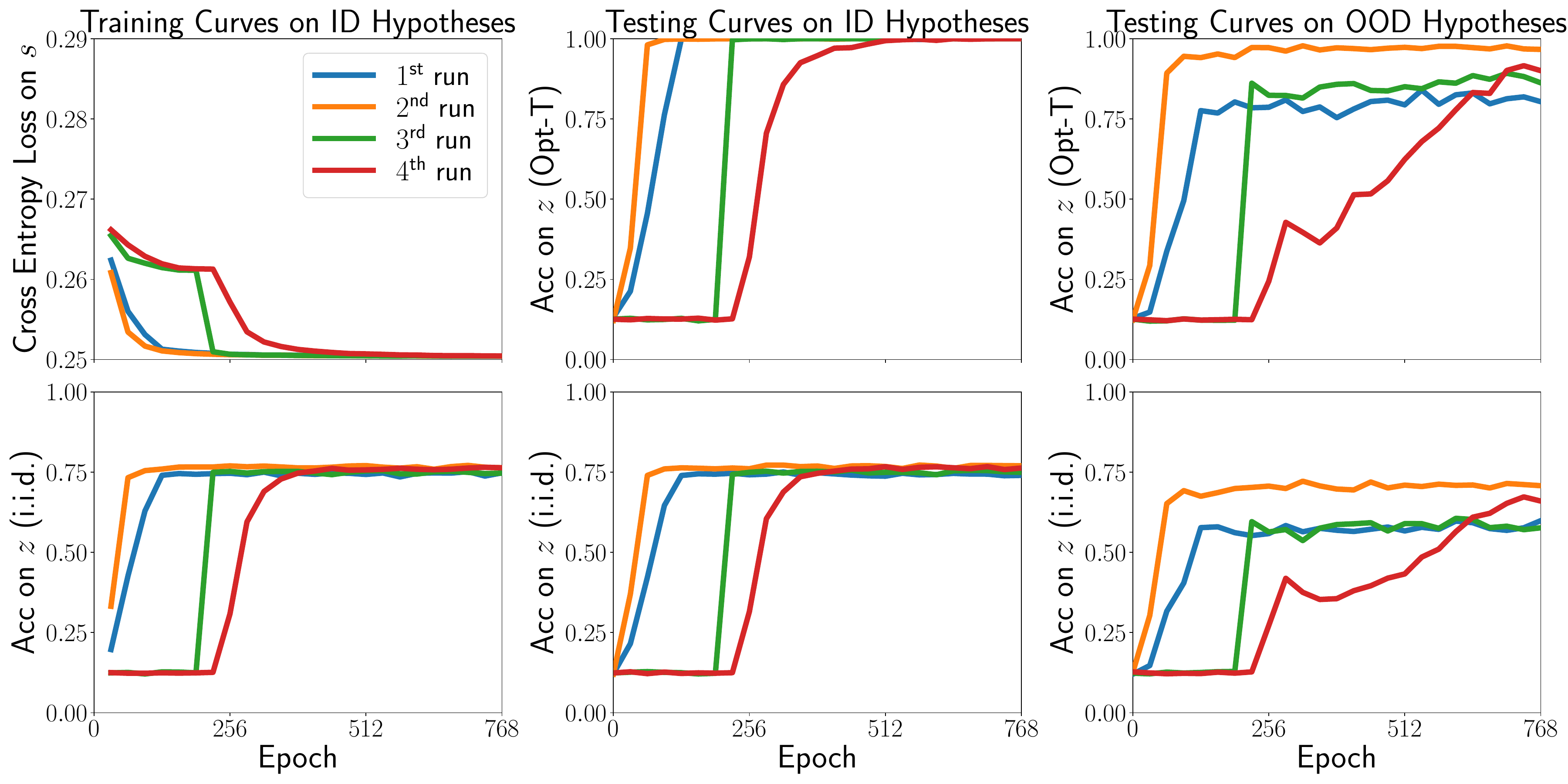}
    \caption{\textbf{Multiple runs for ID and OOD hypothesis class generalizations.}}
    \label{fig:multiple_curves_IO_2x3}
\end{figure*}

\begin{figure*}[th!]
    \centering
    \includegraphics[width = 0.95\textwidth]{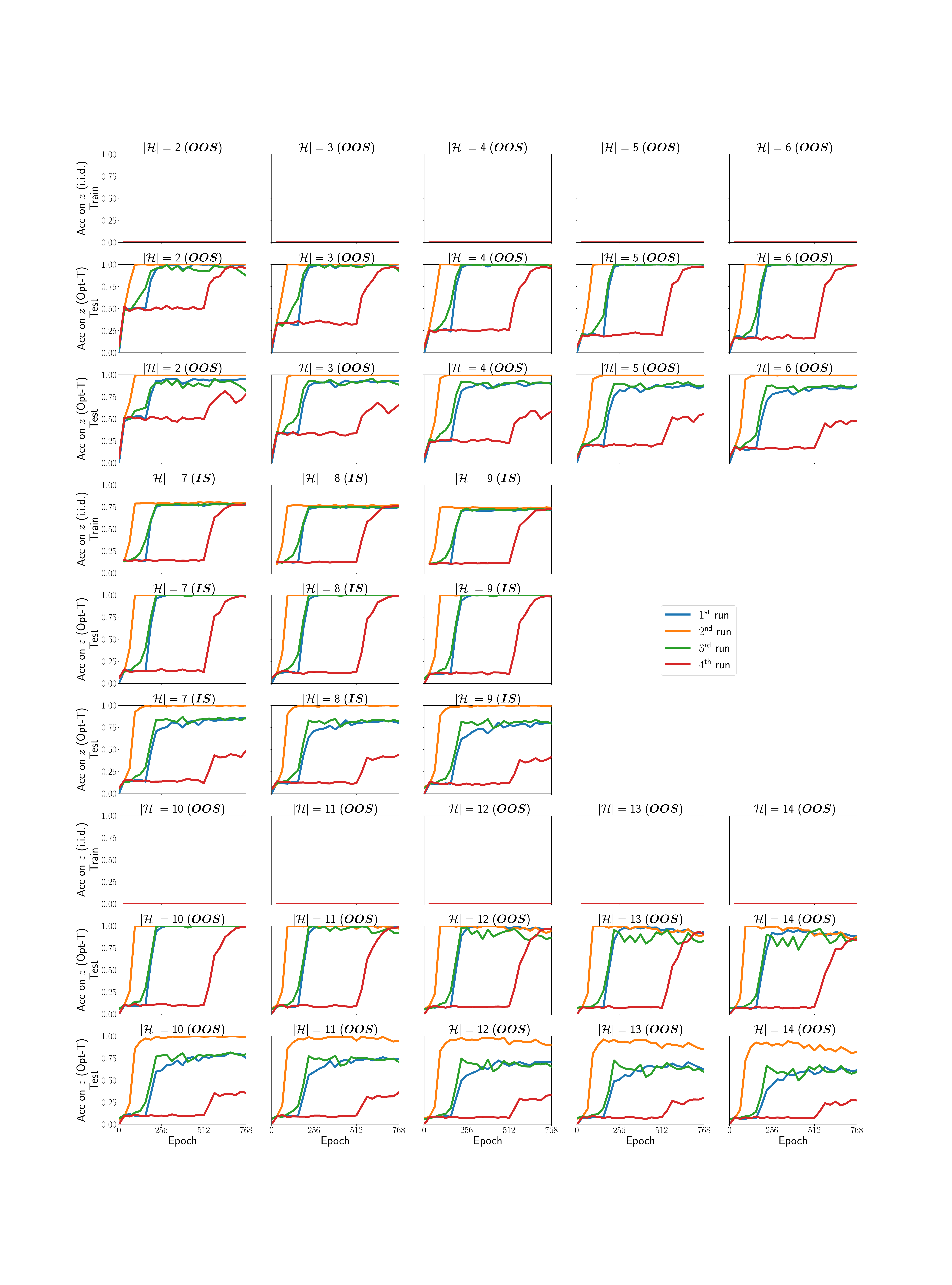}
    \caption{\textbf{Multiple runs for ID and OOD hypothesis class size generalizations.}}
    \label{fig:multiple_curves_IOS_9x5}
\end{figure*}

We share more training and testing curves in Fig.~\ref{fig:multiple_curves_IO_2x3} to provide additional results to Fig.~\ref{fig:multiple_curves_IO}, and in Fig.~\ref{fig:multiple_curves_IOS_9x5} to provide additional results to Fig.~\ref{fig:multiple_curves_IOS}.
\label{subapp:4generalization}

\subsection{Compare with Other Model Architectures}
\label{subapp:4model}
\begin{figure*}[th!]
    \centering
    \includegraphics[width = 0.7\textwidth]{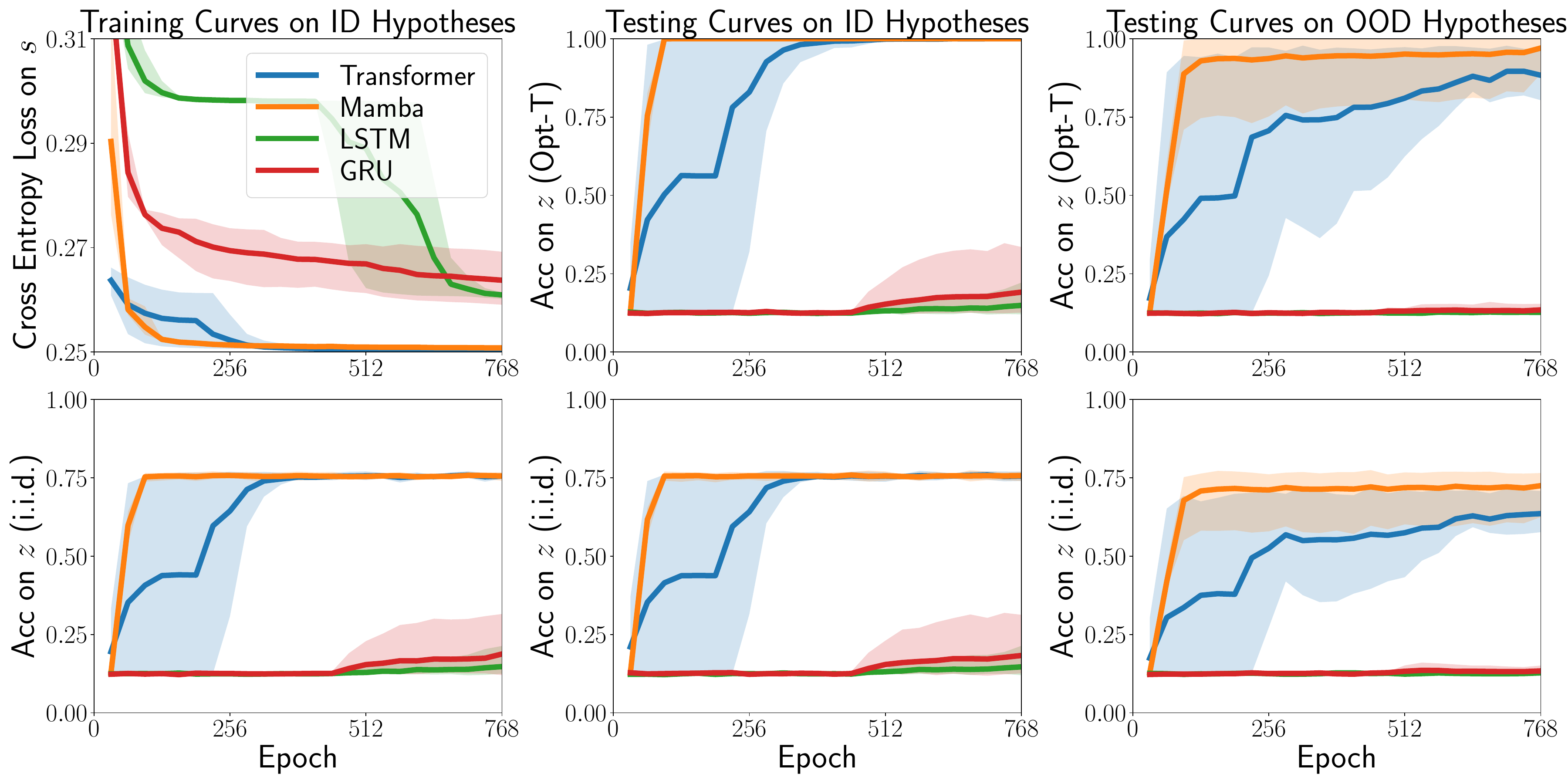}
    \caption{\textbf{Various models on ID and OOD hypothesis class generalizations.}}
    \label{fig:multiple_models_IO_2x3}
\end{figure*}

\begin{figure*}[th!]
    \centering
    \includegraphics[width = 0.95\textwidth]{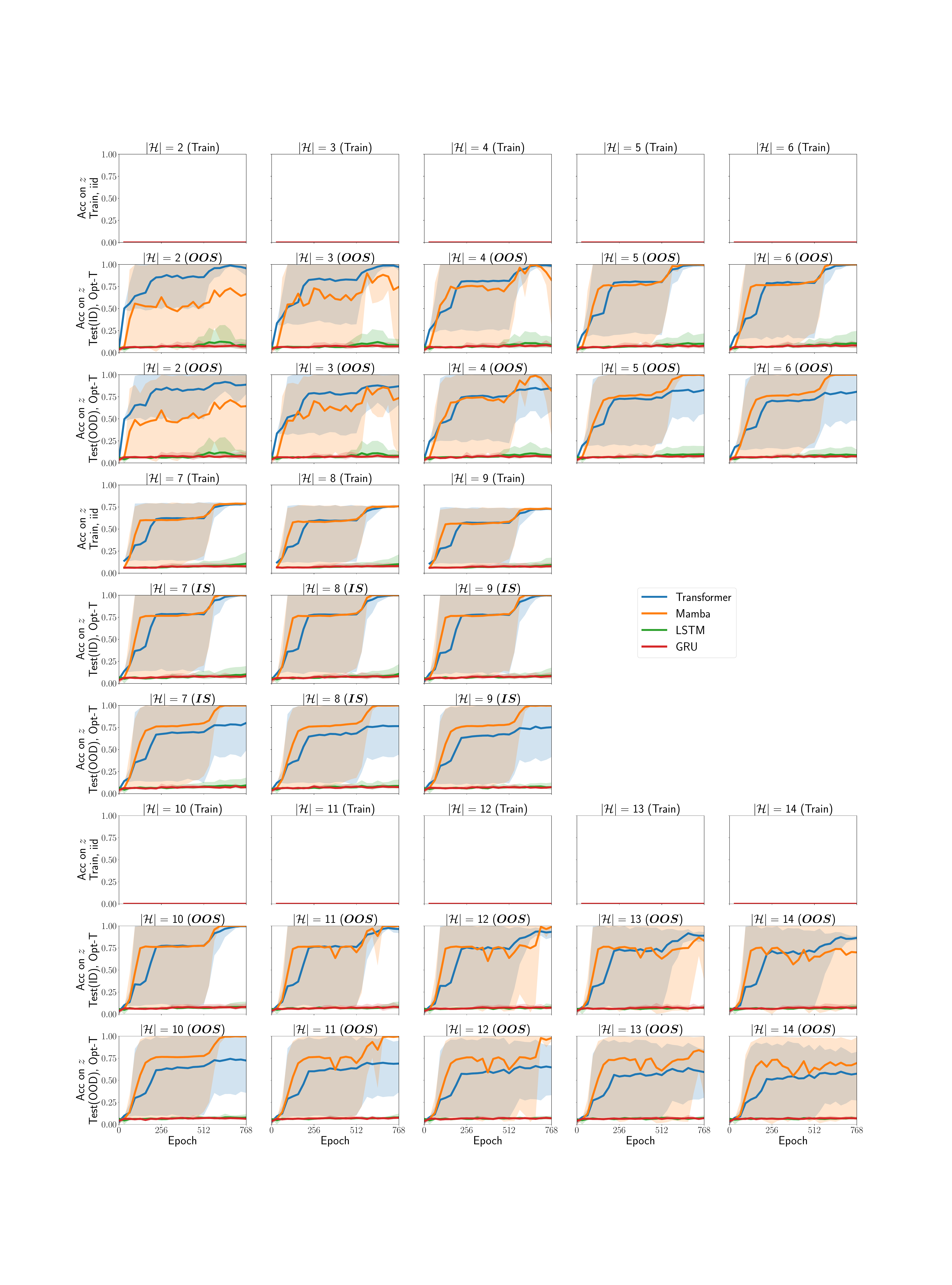}
    \caption{\textbf{Various models on ID and OOD hypothesis class generalizations.}}
    \label{fig:multiple_models_IOS_9x5}
\end{figure*}

We share more training and testing curves in Figs.~\ref{fig:multiple_models_IO_2x3} and~\ref{fig:multiple_models_IOS_9x5} to provide additional results to Figs.~\ref{fig:multiple_models_IO} and~\ref{fig:multiple_models_IOS}, respectively.

\subsection{Effect of Training Class Count}
\begin{figure*}[th!]
    \centering
    \includegraphics[width = 0.9\textwidth]{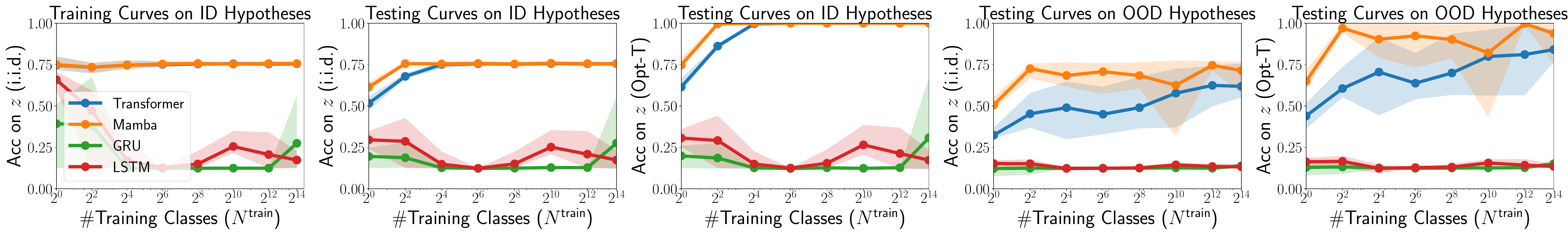}
    \caption{\textbf{Effect of training hypothesis class count on ID and OOD hypothesis class generalization.}}
    \label{fig:num_train_IO_1x5}
\end{figure*}

We share more training and testing curves in Fig.~\ref{fig:num_train_IO_1x5} to provide additional results to Fig.~\ref{fig:num_train_IO}.
\label{subapp:numtrain}

\section{Experiment Setup}
\label{app:expsetup}
Each experiment is \textbf{repeated four times}, with the mean calculated across runs.
The shadow region's boundary is defined by \textbf{the minimum and maximum values} observed across the four runs.
\subsection{Learning Rate Scheduler}
\label{app:lrscheduler}
We set the train procedure with 768 total epochs, each epoch containing 1024 batches.
The learning rate (lr) is first warmed up linearly from an LR$/64$ at epoch $e=1$ to a peak value LR at epoch $e=64$, following:
$$
\text{lr}(e) = \text{LR}\cdot \frac{e}{64}, \quad 1 \leq e \leq 64.
$$
After epoch 64, the learning rate undergoes a quadratic decay over the remaining 704 epochs, given by
\[
\text{lr}(e) = \text{LR}\cdot \sqrt{\frac{64}{e}}, \quad 64 \le e \le 768.
\]

\subsection{Hyperparameter Search}
\label{app:hyperparameters}
We list the hyperparameter searching spaces used for Transformer, LSTM, GRU, and Mamba.
The best hyperparameter is searched using ID hypothesis class generalization with $\|\mathcal{X}\|=5$, $\|\mathcal{H}\|=8$, and then used for all other settings.
\begin{table}[ht!]
\centering
\caption{\textbf{Hyperparamer searching spaces for different model architectures.} The optimal hyperparameters are bolded if multiple possibilities are provided.}
\resizebox{0.98\linewidth}{!}{
\begin{tabular}{lrrrrr}
\toprule
\multicolumn{1}{c}{Model Architecture} & \multicolumn{1}{c}{\#layers} & \multicolumn{1}{c}{\#hidden dimensions} & \multicolumn{1}{c}{\#learning rate} & \multicolumn{1}{c}{\#weight decay} & \multicolumn{1}{c}{\#batch size} \\ \midrule
Transformer                            & 2,\textbf{8}                          & 128                                     & 0.00010, \textbf{0.00020}, 0.00050, 0.00100                             & 0.0005                             & 16                    \\
Mamba                                  & \textbf{2},8                          & 128                                     & 0.00010, 0.00020, \textbf{0.00050}, 0.00100                             & 0.0005                             & 16                    \\
GRU                                    & \textbf{2},8                          & 128                                     & 0.00020, 0.00050, \textbf{0.00100}, 0.00200                             & 0.0005                             & 16                    \\
LSTM                                   & \textbf{2},8                          & 128                                     & 0.00020, 0.00050, \textbf{0.00100}, 0.00200                             & 0.0005                             & 16                    \\ \bottomrule
\end{tabular}
}
\end{table}

\subsection{Setup for Generating Training and Testing Hypothesis Classes}
\label{setup:generalization}
We list the experimental setup for each experiments in the following Table~\ref{table:setup}.
When conducting experiments to evaluate accuracy on $y$, we modified the experimental setup following Table~\ref{table:setupicl}.
\begin{table}[th!]
\centering
\caption{\textbf{Experimental setups of different generalizations.}
The expression \(\min\{512, \#\text{possible}\}\) indicates that when the number of possible hypothesis classes is fewer than 512, we evaluate all possible hypothesis classes for testing; otherwise, we limit the selection to at most 512 hypothesis classes.
For example, if \( |\mathcal{H}^{\text{OOD}}| = 16 \) and \( |\mathcal{H}| = 2 \), the total number of possible hypothesis classes is given by:
$\binom{|\mathcal{H}^{\text{OOD}}|}{|\mathcal{H}|} = \binom{16}{2} = \frac{16 \times 15}{2} = 120.$
Since \( 120 < 512 \), we evaluate all 120 hypothesis classes for testing in this scenario.}
\label{table:setup}
\resizebox{1.00\linewidth}{!}{
\begin{tabular}{lrrrr}
\toprule
\multicolumn{1}{c}{Generalization Setup} & \multicolumn{1}{c}{\begin{tabular}[c]{@{}c@{}}ID Hypothesis\\ Class Generalization\end{tabular}} & \multicolumn{1}{c}{\begin{tabular}[c]{@{}c@{}}OOD Hypothesis\\ Class Generalization\end{tabular}} & \multicolumn{1}{c}{\begin{tabular}[c]{@{}c@{}}ID Hypothesis\\ Class Size Generalization\end{tabular}} & \multicolumn{1}{c}{\begin{tabular}[c]{@{}c@{}}OOD Hypothesis\\ Class Size Generalization\end{tabular}} \\ \midrule
size of input space ($|\mathcal{X}|$)    & 5                                                                                                & 5                                                                                                 & 5                                                                                                     & 5                                                                                                      \\
size of label space ($|\mathcal{Y}|$)    & 2                                                                                                & 2                                                                                                 & 2                                                                                                     & 2                                                                                                      \\
size of context query ($K$)              & 5                                                                                                & 5                                                                                                 & 5                                                                                                     & 5                                                                                                      \\
size of training hypothesis class ($|\mathcal{H}^{\text{train}}|$) & 8                                                                                                & 8                                                                                                 & 7,8,9                                                                                                   & 7,8,9                                                                                                    \\
size of testing hypothesis class ($|\mathcal{H}^{\text{test}}|$)   & 8                                                                                                & 8                                                                                                 & 2$,\ldots,$14                                                                                                  & 2$,\ldots,$14                                                                                                   \\
size of hypothesis prefix ($L$)          & 8                                                                                                & 8                                                                                                 & 16                                                                                                    & 16                                                                                                     \\
\#all hypotheses ($|\mathcal{H}^{\text{uni}}|$)                    & 32                                                                                               & 32                                                                                                & 32                                                                                                    & 32                                                                                                     \\
\#hypotheses in ID pool ($|\mathcal{H}^{\text{ID}}|$)                & 16                                                                                               & 16                                                                                                & 16                                                                                                    & 16                                                                                                     \\
\#hypotheses in OOD pool ($|\mathcal{H}^{\text{OOD}}|$)              & 16                                                                                               & 16                                                                                                & 16                                                                                                    & 16                                                                                                     \\
\#training hypothesis classes            & 12358                                                                                            & 12358                                                                                             & 4096                                                                             & 4096                                                                               \\
\#testing hypothesis classes             & 512                                                                                              & 512                                                                                               & $\min\{512,\#\text{possible}\}$                                                                                & $\min\{512,\#\text{possible}\}$                                                                                 \\ \bottomrule
\end{tabular}
}
\end{table}

\begin{table}[th!]
\centering
\caption{\textbf{Additional setups.} Numbers that differ from those in Table~\ref{table:setup} are highlighted in bold for clarity.}
\label{table:setupicl}
\begin{tabular}{lrr}
\toprule
Section & Sec.~\ref{subsec:icl} & Sec.~\ref{subsec:diversity} \\ \midrule
size of input space ($|\mathcal{X}|$)    & \textbf{4}                                                                                       & \textbf{6}                                                                                        \\
size of label space ($|\mathcal{Y}|$)    & 2                                                                                                & 2                                                                                                 \\
size of context query ($K$)              & \textbf{12}                                                                                      & \textbf{12}                                                                                       \\
size of training hypothesis class ($|\mathcal{H}^{\text{train}}|$) & \textbf{4}                                                                                       & 8                                                                                                   \\
size of testing hypothesis class ($|\mathcal{H}^{\text{test}}|$)   & \textbf{4}                                                                                       & 8                                                                                                   \\
size of hypothesis prefix ($L$)          & \textbf{4}                                                                                       & 8                                                                                                 \\
\#all hypotheses ($|\mathcal{H}^{\text{uni}}|$)                    & \textbf{16}                                                                                & \textbf{64}                                                                                         \\
\#hypotheses in ID pool ($|\mathcal{H}^{\text{ID}}|$)              & \textbf{16}                                                                                & \textbf{8,16,24,32,48}                                                                                                 \\
\#hypotheses in OOD pool ($|\mathcal{H}^{\text{OOD}}|$)            & \textbf{0}                                                                                & 16                                                                                                  \\
\#training hypothesis classes            & \textbf{1308}                                                                                    & $\min\{12358,\#\text{possible}\}$                                                                                            \\
\#testing hypothesis classes             & 512                                                                                              & 512                                                                                               \\ \bottomrule
\end{tabular}
\end{table}

\end{document}